\documentclass{article}

\usepackage{PRIMEarxiv}

\usepackage[utf8]{inputenc} 
\usepackage[T1]{fontenc}    
\usepackage{hyperref}       
\usepackage{url}            
\usepackage{booktabs}       
\usepackage{amsfonts}       
\usepackage{nicefrac}       
\usepackage{microtype}      
\usepackage{lipsum}
\usepackage{fancyhdr}       
\usepackage{graphicx}       
\graphicspath{{media/}}     
\usepackage{mathtools}
\usepackage[bottom]{footmisc}
\DeclareMathOperator*{\argmax}{arg\,max}

\title{Multitemporal SAR images change detection and visualization using RABASAR and simplified GLR
}

\author{
  Weiying Zhao, Henri Maître, Jean-Marie Nicolas and Florence Tupin\\
LTCI, Télécom ParisTech\\
Université Paris-Saclay\\
75013 Paris, France\\
   \And
  Charles-Alban Deledalle \\
  MB, CNRS, Bordeaux INP\\
  University of Bordeaux \\
  F-33405 Talence, France\\
   \And
 Loïc Denis \\
  Laboratoire Hubert Curien UMR 5516, UJM-Saint-Etienne, CNRS \\
  Institut d Optique Graduate School, University of Lyon \\
  F-42023 SAINT-ETIENNE, France\\
}

\begin{document}
\maketitle

\begin{abstract}
Understanding the state of changed areas requires that precise information be given about the changes. Thus, detecting different kinds of changes is important
for land surface monitoring. 
SAR sensors are ideal to fulfil this task, because of their all-time and all-weather capabilities, with good accuracy of the acquisition geometry and without effects of atmospheric constituents for amplitude data. 
In this study, we propose a simplified generalized likelihood ratio ($S_{GLR}$) method  assuming that corresponding temporal pixels have the same equivalent number of looks (ENL). 
Thanks to the denoised data  provided by a ratio-based multitemporal SAR image denoising method (RABASAR), we successfully applied this similarity test approach to compute the change areas. A new change magnitude index method and an improved spectral clustering-based change classification method are also developed.  In
addition, we apply the simplified generalized likelihood ratio  to detect the maximum change magnitude time, and the change starting and ending times. Then, we propose to use an adaptation of the REACTIV method to visualize the detection results vividly. 
The effectiveness of the proposed methods is demonstrated through the processing of simulated and SAR images, and the comparison with classical  techniques. In particular, numerical experiments proved that the developed method has good performances in detecting  farmland  area changes, building area changes, harbour area changes and  flooding area changes.
\end{abstract}


\section{Introduction}

Timely, accurate and continuous monitoring  of land cover and land use changes is  important for  land resource management. 
According to \cite{khorram1999accuracy}, land cover changes can be classified into several categories:
changed from one land cover class to another,    change of shape, shrink or transform,    change of position,    fragment or merge of adjacent regions.
Based on the change reason or change type, changes can be classified into:
 short-term change (synoptic weather events),
    cyclic change (seasonal phenology),
   directional change (urban development),
   multidirectional change (deforestation \& regeneration),
   event change (catastrophic fires).
   Changes happen spatially and temporally. More and more available  multispectral, multitemporal, multisensor satellite data  enhance the capability of detecting, identifying, mapping and monitoring these changes \cite{coppin2004review, habib2007abrupt}.   
   
Optical remote sensing images with high spatial and spectral resolutions can be easily acquired and have been widely used for land cover monitoring  \cite{bruzzone2013novel}. The passive acquisition model of optical sensors and the use of near-visible light wavelengths, sun illumination (or thermal radiation) and cloud-free weather requirements heavily limit its wide application.

However, SAR imagery has the main advantage of being an all-time and all-weather sensor, with good accuracy for the acquisition geometry and with few atmospheric effects for
amplitude data. 
It has been widely used for environmental change detection, urban area change detection and disaster monitoring.
The past few years have seen the launch of numerous synthetic aperture radar (SAR) sensors and new sensors will also be launched soon, such as TSX-NG, Cosmo-SkyMed second generation, Radarsat constellation and NISAR. 
The increasing availability of SAR data allows the high accuracy of change detection, such as abrupt (step) changes, seasonal changes and longer-term developments. 
Unlike optical images, SAR images are seriously affected by speckle noise, which makes the SAR image change detection much harder. 
SAR image change detection can be applied on image pairs or multitemporal series. Different SAR image  characteristics can be used for change detection. 
The current change detection methods are mainly based on likelihood ratio \cite{Lomb-02, su2015norcama,conradsen2003test,conradsen2016determining},  coherence \cite{PCP+16}, image texture and structure analysis \cite{PMM16}, and deep learning based methods \cite{zhao2014deep, gong2016change}. 
Due to the multiplicative noise in
coherent SAR images, the likelihood ratio test is popularly used for change detection. 
A number of studies have applied to SAR data change detection methods based on mean-ratio \cite{TLB88}, image ratio \cite{OQ04, AAD+16}, log-ratio operator \cite{Bazi-05,bovolo2005detail}.
The true distribution of these ratio
images depends on the relative change of the SAR reflectivities \cite{bovolo2015time}.
These methods are easily applied and the associated threshold can be calculated automatically.

Visualizing the changes is also important. Changes usually represent transitions that occur between states \cite{aminikhanghahi2017survey}. 
Using a colourful image can ease results interpretation highlighting changed areas.
 In the change detection field, a number of studies have used different colours to show different changes.  Su et al. \cite{su2015norcama} associate different colours to highlight different change types of time series, Mou et al. \cite{mou2018learning} propose using different colours to represent different change phenomenons,   and Dom\'inguez et al.  \cite{dominguez2018multisquint} use different fusion  strategies to illustrate the real changes and false alarms. With short time series, Nielsen
et al. \cite{nielsen2016change} propose to use RGB colours to represent different change times and use black colour to represent the unchanged areas.  In addition, Amitrano et al.  \cite{amitrano2016end} proposed using new  bi-temporal and multitemporal RGB combination frameworks to illustrate temporal SAR images. The effectiveness of these two RGB visualization approaches has been verified by change detection and classification, respectively.
 However, none of them associates the colours with the times of change in a long time series.
Recently, Rapid and EAsy Change detection in radar TIme-series by Variation coefficient (REACTIV) method has been proposed by Koeniguer et al. \cite{KoeniguerREACTIV}. It is a simple and highly efficient time series change detection and visualization algorithm. 
It is based on HSV  space and exploits only time domain estimates without any spatial estimation. The colour saturation is coded by the temporal coefficient of variation. 
However, the detection results are corrupted by  speckle noise. Even using some state-of-the-art denoising methods, the bias estimation in vegetation areas still prohibits REACTIV to provide accurate performance. In addition, the colour in REACTIV results only represents the appearing date of maximum values.

The  inherent  speckle  which  is  attached  to  any  coherent  imaging  system  affects  the analysis and interpretation of synthetic aperture radar (SAR) images. Thanks to the recently proposed Ratio-Based Multitemporal SAR Images Denoising (RABASAR) method \cite{zhao2019ratio}, we can significantly suppress the negative effect of SAR speckle. 
In this paper,  we derive a simplified generalized log-likelihood ratio criterion ($S_{GLR}$)  based on gamma distribution using RABASAR denoising data. Based on the obtained similarity function,  strategies to apply this similarity criterion to image pair change detection, cumulative change detection and change classification are introduced.
In addition, we adapt REACTIV method to integrate RABASAR denoising results. Then, we apply the simplified GLR function in this framework to detect different change times of interest, such as change starting and ending times under predefined threshold, or maximum change magnitude appearing time.

\section{Image pair change area detection and change magnitude visualization}

In this section,  we derive a generalized log-likelihood ratio criterion based on gamma distribution using denoised data. Based on the obtained similarity functions,  strategies to apply this similarity criterion to image pair change detection, cumulative change detection, change classification and change time detection are introduced.

\subsection{Change area detection}

Under Goodman's hypothesis \cite{goodman2007speckle}, the fully developed intensity speckle follows a Gamma distribution $\mathcal{G}[u,L]$ depending on the number of looks $L$ and the mean reflectivity $u$ of the scene:
\begin{equation}
    \mathcal{G}[u,L](y)=\frac{L}{u\Gamma(L)}\bigg(\frac{Ly}{u}\bigg)^{L-1} e^{-\frac{Ly}{u}}
    \label{eq:Gamma}
\end{equation}

To compare the similarity of two gamma distributed variables $(y_t, y_{t'})$, a log-likelihood ratio test can be used. Speckle is an inherent problem for SAR image interpretation which brings many drawbacks for traditional SAR image change detection methods. 
When dealing with denoised images $\hat{u}_{t}$ and $\hat{u}_{t'}$, with associated ENL $L_t$ and $L_{t'}$, we have:
\begin{equation}
    S_{GLR}(\hat{u}_t,\hat{u}_{t'})=
    L_t\log \frac{L_t \hat{u}_t+L_{t'}\hat{u}_{t'}}{\hat{u}_t(L_t+L_{t'})} + L_{t'}\log \frac{L_t \hat{u}_t+L_{t'}\hat{u}_{t'}}{\hat{u}_{t'}(L_t+L_{t'})}
\end{equation}
where $\hat{u}$ is the estimated reflectivity, $t$ represents the time index in the time series. 

Unlike $CGLRT$ and $AGLRT$ methods \cite{su2015norcama}, we fully trust the denoising results and do not take the noisy data into account any more. In practice, we directly calculate the similarity of the multi-looked SAR data with the use of their corresponding ENL. 
Suppose the corresponding pixels in  despeckling images have the same ENL $L_t=L_{t'}=\hat{L}$,  the simplified GLR  method \cite{deledalle2009iterative,SDT+14} turns out to be: 
\begin{equation}\label{eq:glr}
    S_{GLR}(\hat{u}_t,\hat{u}_{t'})=2\hat{L}\log{\bigg(\sqrt{\frac{\hat{u}_t}{\hat{u}_{t'}}}+
    \sqrt{\frac{\hat{u}_{t'}}{\hat{u}_t}}}\bigg) -2\hat{L}\log{2}
\end{equation}

Defining a global threshold is a simple and widely used approach to distinguish changes from unchanged points.
To detect the changed areas, we used a thresholding function:
\begin{equation}
    \varphi[S_{GLR}(\hat{u}_t,\hat{u}_{t'})] =
    \left\{
        \begin{array}{cc}
                1, & \mathrm{if\ } S_{GLR}(\hat{u}_t,\hat{u}_{t'}) \geq \tau \\
                0, & \mathrm{otherwise} \\
        \end{array} 
    \right.
\end{equation}

The threshold definition methods are under the same no-change hypothesis framework that can be used for this method.
As introduced by  \cite{wilks1938large}, when the sample size approaches infinity, the log-likelihood statistic model asymptotically converges towards chi-squared distributed probability under the null hypothesis. Thus, we can use the chi-square cumulative function to estimate the change probability of  $S_{GLR}(\hat{u}_t,\hat{u}_{t'})$ with:
     \begin{equation}
     \rho=1-\frac{1}{4\hat{L}}
 \end{equation}
 \begin{equation}
     \omega_2=-\frac{1}{4}(1-\frac{1}{\rho})^2
 \end{equation}
\begin{equation}
    P\{2\rho S_{GLR}(\hat{u}_t,\hat{u}_{t'})\le \delta\} \simeq P\{\chi^2 (1)\le \delta\}+\omega_2[P\{\chi^2 (5)\le \delta\}-P\{\chi^2 (1)\le \delta\}]
\end{equation}
where $\delta$ is the statistical significance. For the detailed derivation of this probability calculation method, we recommend referring to \cite{conradsen2003test, anderson1958introduction}. Unlike Conradsen's PolSAR change detection analysis, we mainly pay attention to spatially adaptive denoised single-channel SAR images with the same estimated ENL.  To robustly estimate $\hat{L}$ in the denoised image, the log-cumulant method is used \cite{tison2004new}.  In the following sections, we mainly use this way to define the threshold.

\subsection{ Change magnitude index for visualization}

To distinguish  appearing from disappearing changes, we used a signum function $\texttt{sign}(x)$ to convert $S_{GLR}(\hat{u}_t,\hat{u}_{t'})$ to  positive and negative values. In this case, if we set the image acquired at time $t$ as the reference image, the positive and negative values correspond to the increase and decrease of the object backscattering values.

\begin{equation}
    \texttt{sign}(x) 
    \begin{cases}
      -1 & \texttt{if } x < 0 \\
      0  & \texttt{if } x = 0 \\
      1  & \texttt{if } x > 0
    \end{cases}
  \end{equation}

\begin{equation}
    x=\log \bigg(\sqrt{\frac{\hat{u}_{t'}}{\hat{u}_{t}}}\bigg)
\end{equation}

To clearly illustrate the temporal changes, the similarity ratio is normalized and transformed to values within the range [0, 255]. 

\begin{equation}
    S^{conv}_{GLR}(\hat{u}_t,\hat{u}_{t'}) =
    \left\{
        \begin{array}{cc}
                255, & \mathrm{if\ } \frac{2(S_{GLR}(\hat{u}_t,\hat{u}_{t'})-\alpha_1)}{\alpha_2-\alpha_1} \geq 2 \\
                127\frac{2(S_{GLR}(\hat{u}_t,\hat{u}_{t'})-\alpha_1)}{\alpha_2-\alpha_1}+1, & \mathrm{otherwise} \\
        \end{array} 
    \right.
\end{equation}
where $\alpha_1$ and $\alpha_2$ represent the minimum and maximum values in the temporal dissimilarities $S_{GLR}(\hat{u}_t,\hat{u}_{t'})$. To suppress the outliers, we empirically set $\alpha_1=-2$ and $\alpha_2=2$.
The value range will be converted to [-255, 255] by multiplying $S^{conv}_{GLR}(\hat{u}_t,\hat{u}_{t'})$ with $\texttt{sign}(x)$.
The rainbow index colour is used to represent different change magnitudes (appearing, disappearing and slow changes).

In practice, we can arbitrarily combine the reference and slave image through RGB composition (R: slave image, G: reference image, B: slave image). Although different colours could indicate the increase and decrease of backscattering values, the illustration performance is not as good as the aforementioned strategy.

\subsection{Time series change type classification}
During the time series acquisition, changes may occur multiple times and with different magnitudes. To detect the change types, we propose an improved change classification method inspired by NORCAMA method  \cite{su2015norcama} and spectral clustering method \cite{shi2000normalized, ng2002spectral}. 
In practice, the change types  are transferred into a partitioning problem and detected using  spectral clustering.

\subsubsection{Change Criterion Matrix (CCM)} 

By making use of the eigenvalues of the similarity matrix of the data, spectral clustering techniques perform dimensionality  reduction before clustering in fewer dimensions. It has been successfully used to cluster the temporal pixels based on their similarity symmetric matrix \cite{ng2002spectral}.
Given a time series $\{\hat{u}_1, \hat{u}_2\cdots \hat{u}_M\}$, the affinity matrix is defined as a symmetric matrix $A(s)$, with elements $S(\hat{u}_t,\hat{u}_{t'}) $ representing the change criterion between different data points. 
\begin{equation}\label{eq:changeCriterionMatix}
A(s)=
\begin{bmatrix}
    S(\hat{u}_1,\hat{u}_1) & S(\hat{u}_1,\hat{u}_2) & S(\hat{u}_1,\hat{u}_3) & \dots  & S(\hat{u}_1,\hat{u}_{t'}) \\
    S(\hat{u}_2,\hat{u}_1) & S(\hat{u}_2,\hat{u}_2) & S(\hat{u}_2,\hat{u}_3) & \dots  & S(\hat{u}_2,\hat{u}_{t'}) \\
    \vdots & \vdots & \vdots & \ddots & \vdots \\
    S(\hat{u}_t,\hat{u}_1) & S(\hat{u}_t,\hat{u}_2) & S(\hat{u}_t,\hat{u}_3) & \dots  & S(\hat{u}_M,\hat{u}_M)
\end{bmatrix}
\end{equation}
where $A(s)$ is a symmetric change criterion matrix with size $M \times M$, $s$ is the location in one image, $t$ and ${t'}$ are time index with $1 \le t\le M$ and $1 \le {t'}\le M$.

To avoid the overlapping of different clusters, Xin $et$ $al.$ \cite{su2015norcama} proposed to binarize the change criterion matrix $A(s)$. The binary process can tighten the clusters, but it will force the clustering results seriously depending on the used thresholds. 
In practice, we use a binarized change criterion matrix. In addition, the $k$-nearest neighbours algorithm could be used to classify this  change criterion matrix as well.
To suppress the temporal variance caused by the residual speckle, we can apply the exponentially weighted moving average \cite{schubert2014signitrend} to the time series.

\subsubsection{Clustering by spectral clustering method}

Based on the acquired change criterion matrix, the Laplacian matrix $A^L(s)$ is computed by:
\begin{equation}
    A^L(s) = D(s) - A(s)
\end{equation}

\begin{equation}
D(s)=
\begin{bmatrix}
    \sum S(\hat{u}_1,\hat{u}_{t'}) & 0 &  \dots  & 0 \\
    0 & \sum S(\hat{u}_2,\hat{u}_{t'}) &  \dots  & 0 \\
    \vdots & \vdots  & \ddots & \vdots \\
    0 & 0 & \dots  & \sum S(\hat{u}_t,\hat{u}_{t'})
\end{bmatrix}
\end{equation}

\begin{equation}\label{eq:diagnalCalcualtion}
    \sum S(\hat{u}_t,\hat{u}_t)=\sum_{{t'}=1}^{M} S(\hat{u}_t,\hat{u}_{t'})
\end{equation}

The former steps (Eq.(\ref{eq:changeCriterionMatix})$\sim$(\ref{eq:diagnalCalcualtion})) are the same as NORCAMA \cite{su2015norcama} except the similarity calculation. Then, the Laplacian matrix $A^L(s)$ is normalized using:
\begin{equation}\label{eq:normLaplacian}
    A^L_{norm}={D}^{-1/2}A^L{D}^{-1/2}
\end{equation}
The eigenvalues $\lambda$ are computed through: 
\begin{equation}
     A^L_{norm}V=\lambda V
\end{equation}
where $V$ is the eigenvector.
After sorting the acquired eigenvalues $\{\lambda_1, \lambda_2, \cdots, \lambda_M \}$ in ascending order, the clustering number $k$ is calculated using the eigengap heuristic method \cite{von2007tutorial}:
\begin{equation}
    k=\argmax_{1 \le t <M}(\lambda_{t+1} - \lambda_t)
\end{equation}

To reduce the data dimension, only the eigenvectors $\upsilon_t$ ( $M\times 1$ column 
vector) which correspond to the $k$ largest eigenvalues of $L^{norm}$ are used with $U=[\upsilon_1,\upsilon_2\cdots\upsilon_k]$. 
To obtain the unit norm, we re-normalize the matrix rows. 
Finally, the $k$-means method is used to  cluster each row $u_t$ in $U$ and the cluster labels $\{l_{1},l_{2},... l_{M}\}$ are assigned to each cluster element with $1 \le l_{t}\le k$.

The aforementioned method is similar to the normalized cut method  \cite{shi2000normalized, su2015norcama}. However, they normalize the rows of $A(s)$ to sum to 1 and use its eigenvectors instead of the normalized Laplacian matrix $A^L(s)$ calculated using equation (\ref{eq:normLaplacian}). In addition, they do not re-normalize the rows of $U$ to unit length \cite{ng2002spectral}.

\subsubsection{Change type recognition} 

Based on the number of clusters $k$ and cluster labels \{$l_{1},l_{2},... l_{M}$\} acquired by the $k$-means algorithm, the change type of the time series points can be recognized \cite{su2015norcama} according to Table \ref{changeType}.

\begin{table}
\centering
\caption{Label of different change types}
\label{changeType}
\begin{tabular}{cccc}
\toprule
\textbf{Classes}&\textbf{Types} & \textbf{$k$} & \textbf{Label series \{$l_{1},l_{2},... l_{M}$\}} \\
\midrule
1&Unchanged & 1 & {1, 1, ...1}\\
2&Step & 2  &{1, 1, ...1, 2, 2, ...2}\\
3&Impulse & 2  &{1, 1, ...1, 2, 2, ...2, 1, 1, ...1}\\
4&Cycle & 2  &{1, ...1, 2, ...2, 1, ...1, 2,...2,...}\\
5&Complex & $\geq$3  & {1, 1, ..., 2, 2..., 3, 3...4, 4...}\\
\bottomrule
\end{tabular}
\end{table}

\section{Change times of interest detection and visualization with extended REACTIV}

In this section, our objective is to adapt the REACTIV method to integrate RABASAR denoising results. Then, using the proposed simplified GLR function to detect the change time of interest.

The principle of REACTIV is to exploit the HSV colour space and a temporal stack of SAR images. The hue channel H represents the time,
the saturation channel S corresponds to the temporal coefficient of variation, and the value V corresponds to the maximum radar intensity of the temporal series in each pixel \cite{koeniguervisualisation}. 

\subsection{Times of interest (Hue)}\label{se:ChangeTime}

As introduced in \cite{koeniguervisualisation}, one can associate a colour with a particular time according to the change. During the procedure, different change types can be considered, such as abrupt change, seasonal change, deforestation \& regeneration, etc. The REACTIV visualization method chooses to highlight the  appearing time of the maximum value.
 Although the REACTIV visualization method can well associate the maximum  value appearing time with the colour, the first and last dates have very similar colours because the HSV colour palette is continuous and loops on itself.
Thus, we propose to highlight the interested time using the normalized time in the interval:

\begin{equation}\label{eq:hue}
    f_t=\frac{5}{6}\times\frac{t-t_1}{t_2-t_1}
\end{equation}
where $t_1$ and $t_2$ are the first and last image acquisition time in the time series, and $t$ is the time of interest. ${5}/{6}$ is used to suppress the time interval, so as to avoid using the starting color category (loop) again.

With the time series $\{\hat{u}_1(s), \hat{u}_2(s), \cdots, \hat{u}_M(s)\}$, we can use $S_{GLR}$ to detect times of interest: 

\begin{itemize}
\item Start changing time

    When detecting the start changing time in the time series, points similarities $S_{GLR}(\hat{u}_1(s),\hat{u}_{t}(s))$ are calculated with the reference to the value of the first date $\hat{u}_1(s)$. After transforming the similarity to change probability $P\{2\rho S_{GLR}(\hat{u}_1,\hat{u}_{t'})\}$, we can decide whether there is a change or not based on a predefined change probability $\tau$ (such as $99\%$). 
    
    \begin{equation}
    T_{start}=
    \begin{cases}
      t & \texttt{if } P\{2\rho S_{GLR}(\hat{u}_1,\hat{u}_{t'})\} > \tau \\
      0  & \texttt{else}
    \end{cases}
  \end{equation}
  where $t$ is the start changing time with $1<t\le M$. It corresponds to the first time that the change probability is larger than the threshold $\tau$.

\item {Maximum changing time}

    Generally, the abrupt changes are associated  with large change magnitude \cite{basseville1993detection}. The  $S_{GLR}(\hat{u}_t(s),\hat{u}_{t'}(s))$  is supposed to be the maximum change. In this case, $t$ and $t'$ are set as the adjacent times. $t'$ is the  maximum change time.
    
\begin{equation}
    T_{max}= 
    \begin{cases}
      t' &  \text{if } S_{GLR}(\hat{u}_t(s),\hat{u}_{t'}(s))=  \tau_{max}\\
      0  & \text{else}
    \end{cases}
\end{equation}
  where $\tau_{max}$ equals to the maximum dissimilarity $\tau_{max}=\argmax\limits_{t'}S_{GLR}(\hat{u}_t(s),\hat{u}_{t'}(s))$.

\item {Stop changing time}      

    For the detection of stop changing time, the last changed point in the time series is used as the reference point. In practice, the calculation is carried out in reverse order.
    
  \begin{equation}
    T_{stop}= 
    \begin{cases}
      t & \text{if } P\{2\rho  (S_{GLR}(\hat{u}_t(s),\hat{u}_M(s)))\} > \tau \\
      0  & \text{else}
    \end{cases}
  \end{equation}
  where $t$ corresponds to the dates going from $M, M-1, \cdots$ to 1.

\end{itemize}  

\subsection{Saturation (S)}  

This colour component defines whether there are changes or not in the time series. Unlike popularly used change detection methods which mainly pay attention to the intensity value changes, the REACTIV method uses the dynamics of the coefficient of variation. 
Based on Rayleigh Nakagami's distribution \cite{G76}, it is possible to derive the empirical moments' expression of pure speckle with \cite{nicolas2006application}:

\begin{equation}
    m_1=u_A\frac{\Gamma(L+\frac{1}{2})}{\sqrt{L}\Gamma(L)}
\end{equation}
\begin{equation}
    m_2=u_A^2
\end{equation}

Based on the ratio of standard deviation and the amplitude average, the coefficient of variation can be calculated through:

\begin{equation}
    \gamma=\frac{\sigma}{u_A}=\frac{\sqrt{m_2-m_1^2}}{m_1}=\sqrt{\frac{\Gamma(L)\Gamma(L+1)}{\Gamma(L+1/2)^2}-1}
\end{equation}

To know the behaviour of this parameter,  the variance of this estimator can be calculated according to \cite{nicolas2006application, kendall1977advanced}:

\begin{equation}
   \mathrm{Var}(\gamma)=\frac{1}{4M}\frac{4m_2^3-m_2^2m_1^2+m_1^2m_4-4m_1m_2m_3}{m_1^4(m_2-m_1^2)}
\end{equation}
where $M$ is the number of samples to compute the estimation which corresponds here to the temporal images considered. 
In addition, based on the order 1 to 4 moments of the Nakagami distribution and $L$, the variance of this estimator can be written:

\begin{equation}
    \mathrm{Var}(\gamma)=\frac{1}{4M}\frac{L\Gamma(L)^4(4L^2\Gamma(L)^2-4L\Gamma(L+\frac{1}{2})^2-\Gamma(L+\frac{1}{2})^2)}{\Gamma(L+\frac{1}{2})^4(L\Gamma(L)^2-\Gamma(L+\frac{1}{2})^2)}
\end{equation}

Usually, the number of available images on the same area varies. In order to overcome the dependency of the coefficient of variation on the number of images, it is possible to normalize the distribution with the  theoretical mean and standard deviation. Koeniguer
et al. \cite{koeniguervisualisation} propose to use the following empirical normalization:

\begin{equation}
    \gamma \leftarrow \frac{\gamma - \mathbb{E}[\gamma]}{10\sigma(\gamma)}+0.25
\end{equation}
with the theoretical mean and standard deviation values for a "stable" speckle and $L = 4.9$ for Sentinel-1 GRD data. This empirical normalization aims to reduce the saturation values of the stable zones around a low saturation to 0.25 and to spread the changes on higher saturation.

To speed up the processing, we will use  a coefficient of variation method to detect time series changes. Then, detect the different change times of interest.
    
\subsection{Value (V)} 

The REACTIV visualization method uses the maximum amplitude value of each time series as the value channel. The hue colour component is computed using equation (\ref{eq:hue}) with time $t$ corresponding to the maximum amplitude value appearing time.  This choice is particularly suitable for an abrupt event (such as the presence of a boat). 

Although using maximum values can highlight abruptly appearing objects, the acquired results seem too noisy.  Apart from this choice, we can use the denoising results or the temporal average image to have a clear vision of the ground. We only use the maximum time series values in the following experiments.

\section{Experimental results and discussion}

To illustrate and compare the proposed methods with state-of-the-art change detection methods,  simulated SAR images and  SAR images are tested in this section. All the data are despeckled by RABASAR before change analysis: 
image pair change detection in section \ref{se:imagePairComparison}, continuous change monitoring in section \ref{se:cumulativeChangeDetection}, change classification experiments in section \ref{se:changeClassification} and change time detection in section \ref{se:changeTime}.

\subsection{Experimental data introduction}

To test the improved algorithms, we prepared four kinds of data. In this paper, we mainly consider SAR images acquired through the same orbit with similar incidence angles. During the RABASAR denoising, we mainly use arithmetic mean (AM), denoised arithmetic mean (DAM) and denoised binary weighted arithmetic mean (DBWAM) of the time series.

\begin{itemize}
    \item Simulated SAR data
    
    Many SAR image simulations are based on reflectivity maps obtained from optical images. However, real SAR images exhibit strong and persistent scatterers, especially in urban areas which can hardly be simulated using optical images.
Therefore, we propose to use the arithmetic mean image of  long time series of SAR images, considered as a noise-free image (a reflectivity map $u$) to create realistic simulations of SAR images. With the acquired arithmetic mean image $y^{AM}$, we add different kinds of object changes according to their values in real SAR images. Then, we simulate the temporal images by multiplying $y^{AM}$ with different simulated Gamma distribution noise $v_t$.  
    
    \item TerreSAR-X data
    
The TerraSAR-X images are acquired over the harbour area of $Sendai$.
9 temporal well-registered SAR images are used for the preparation of denoised data, both for 2SPPB \cite{SDT+14}  and for RABASAR.  Only the two images which were acquired on 06/05/2011 and  08/06/2011 are used for the change detection. 

    \item Sentinel-1 single look SAR data
    
    The used Sentinel-1 IW VV single look SAR data are acquired over the Saclay area, South of Paris. All the images are registered using geometric-based subpixel images registration method \cite{JERFFO12}. 
    This area mainly contains farmland area, forest area, building area, etc.
    Saclay area has been chosen since it has been chosen to receive the future scientific area of the University of Paris-Saclay. Starting in 2010, constructions and public works were decided to convert agricultural terrains into research and education buildings, mostly 2 to 5 storey compact and geometrical structures made of concrete, steel and glass. Many plots have been barred from vegetation, and the excavated heavy plant machinery and trucks have been parked in some places. All these elements greatly influence the SAR reflectivity.
    
    \item Sentinel-1 GRD data
    
    The Sentinel-1 GRD data are acquired over Saddle Dam D, Southern Laos. All the images are coregistered using geometric-based registration method\footnote{SNAP: http://step.esa.int/main/toolboxes/snap/}. All the historical Sentinel-1 CRD VV descending images acquired over this area are used to compute the temporal mean image, which is used by the RABASAR method for image denoising. Google Engine Engine is used during the image preparation.
    
\end{itemize}

\subsection{Image pair change detection}\label{se:imagePairComparison}

To evaluate the change detection performances and validate the effectiveness of  $S_{GLR}$ method, the image pair change detection results are compared with Conradsen's method \cite{conradsen2003test, conradsen2016determining} and $CGLRT$ method \cite{su2015norcama}.

\begin{figure}
\graphicspath{{results/}}
   \centering
\begin{tabular}{cccc}
\multicolumn{2}{c}{\includegraphics[width=4.4cm]{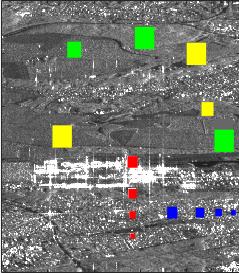}}&
\multicolumn{2}{c}{\includegraphics[width=4.4cm]{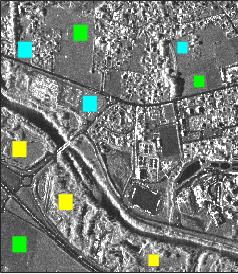}}\\
 \multicolumn{2}{c}{(a)}& \multicolumn{2}{c}{(b)}\\
\multicolumn{2}{c}{\includegraphics[width=7cm]{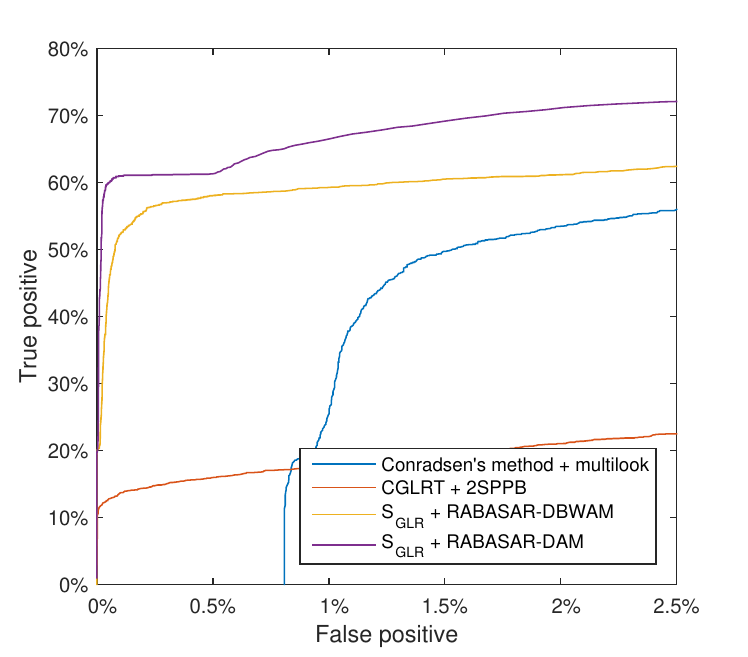}}&
\multicolumn{2}{c}{\includegraphics[width=7cm]{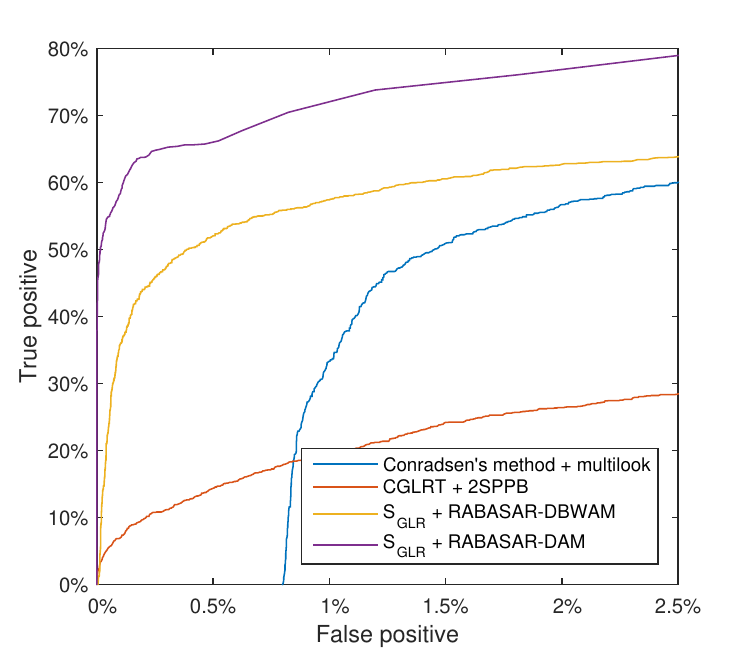}}\\
 \multicolumn{2}{c}{(c)}& \multicolumn{2}{c}{(d)}\\
\end{tabular}
  \caption{False positive vs true positive curves comparison based on simulated SAR images. (c) ROC curve results using simulated Sentinel-1 data (a), (d) ROC curve results using simulated TerraSAR-X data (b).
  Multilooked data with $3\times3$ window size, 2SPPB and RABASAR provided data are used for the comparison of Conradsen's method, $CGLRT$ method and $S_{GLR}$ method, respectively. Different colours represent different object changes: green=farmland, yellow=forest, red=appearing, blue=appearing then disappearing, cyan=disappearing.}
  \label{fig:ROCImgPairComparison} 
\end{figure}

During the acquisition of SAR time series, changes may happen for different kinds of objects, such as buildings, farmland  and forest, etc. Generally, different kinds of objects have different change magnitudes.
To comprehensively and quantitatively evaluate the performances of different methods, we processed the simulated SAR images which have different kinds of object changes.

Because of the small change magnitude in SAR intensity images, identifying forest area changes is much harder than farmland  and building  changes. It is obvious that $S_{GLR}$ method can obtain the best detection results. In addition, $S_{GLR}$ method can obtain better results with RABASAR-DAM provided data.
Compared with other methods, the $CGLRT$ method is good at detecting building area changes which have large change magnitude. This characteristic causes  $CGLRT$ method to  provide worse results when the false positives are larger than $1\%$. According to the results shown in Figure \ref{fig:ROCImgPairComparison}, Conradsen's method  always have more false positive because of using multilooked data.

\begin{figure}
\graphicspath{{results/}}
   \centering
\begin{tabular}{cc}
\includegraphics[width=7cm]{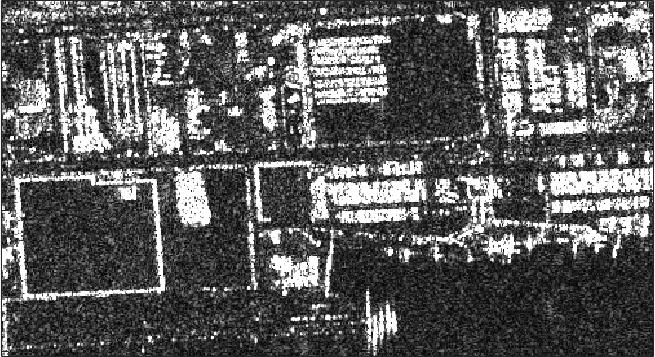}&
\includegraphics[width=7cm]{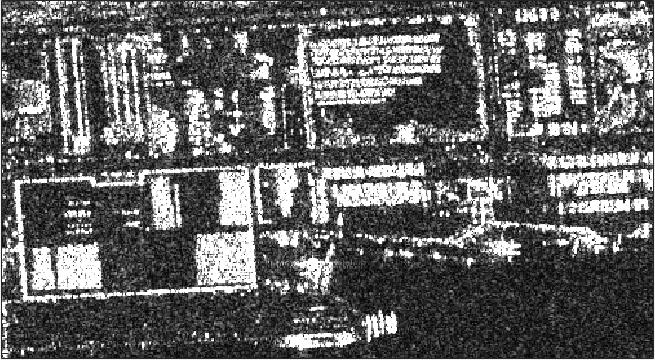}\\
 (a) Noisy image acquired on 06/05/2011&(b) Noisy image acquired on  08/06/2011\\
\includegraphics[width=7cm]{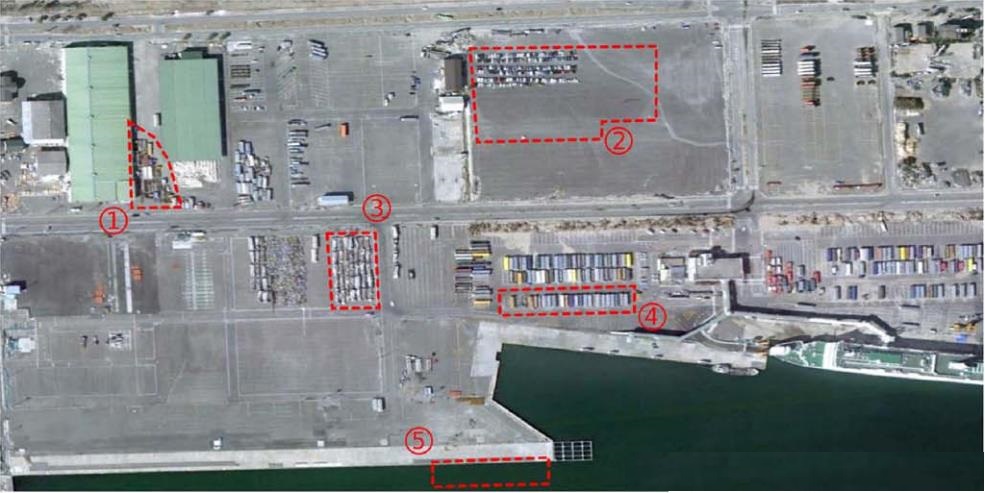}&
\includegraphics[width=7cm]{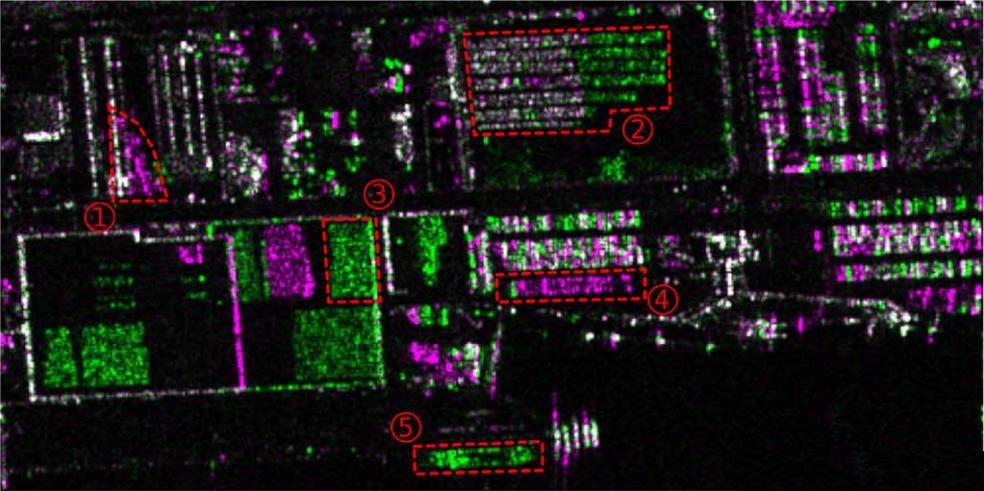}\\
 (c) Optical image acquired on 06/04/2011&(d) RGB combination of the noisy images\\
\includegraphics[width=7cm]{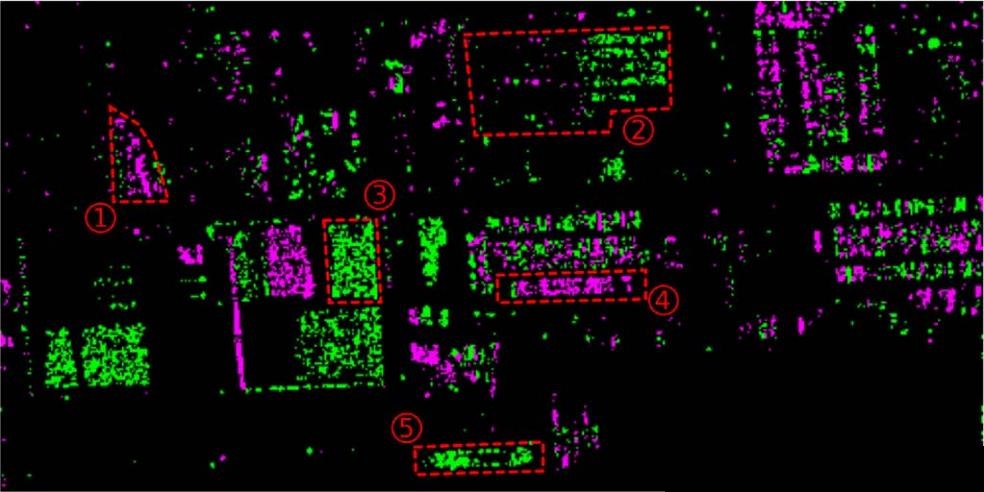}&
 \includegraphics[width=7cm]{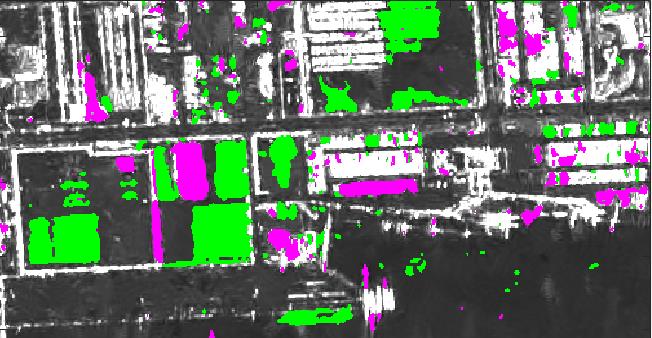}\\
 (e) MIMOSA with false alarm rate equal to $1\%$&(f) $CGLRT$  with 2SPPB  data\\
\includegraphics[width=7cm]{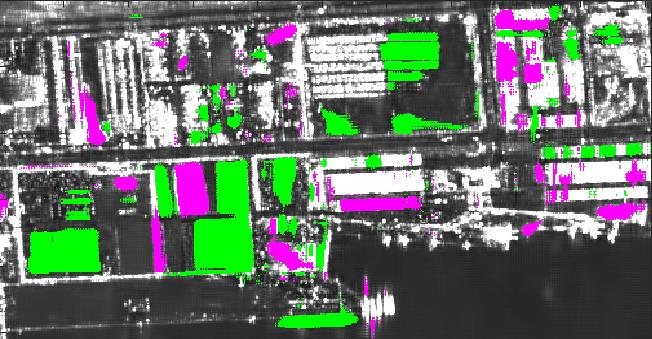}&
\includegraphics[width=7cm]{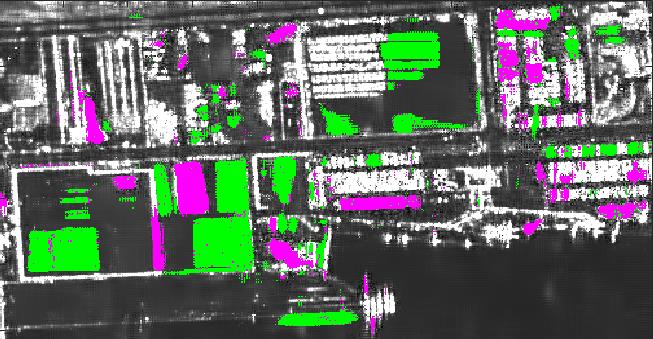}\\
 (g) $S_{GLR}$  with RABASAR-DAM  data&(h) $S_{GLR}$ with RABASAR-DBWAM  data\\
\end{tabular}
  \caption{ $Sendai$ SAR image pair change detection comparison. Pink represents disappearing areas, while green represents appearing areas. The comparison areas are indexed using red circles with associated numbers.}
  \label{fig:sendaiImgPairComparison} 
\end{figure}

    To fairly compare with the state-of-the-art change detection methods, we processed  popularly used  TerraSAR-X images  acquired over $Sendai$. 
Although MIMOSA \cite{quin2014mimosa} detects all the changes directly using the noisy data, there are too much wrong detections in the unchanged areas. The detections are seriously influenced by the noise.
$CGLRT$ can provide good results, but the changed area boundaries are blurred compared with $S_{GLR}$. In addition, $CGLRT$ provides some wrong results in the water area.

\begin{figure}
\setlength{\abovecaptionskip}{-10pt} 
\setlength{\belowcaptionskip}{-10pt} 
\graphicspath{{results/}}
   \centering
\begin{tabular}{cccc}
\footnotesize{\rotatebox[origin=l]{90}{ 24/12/2014-05/05/2015}}&
\includegraphics[width=4.4cm]{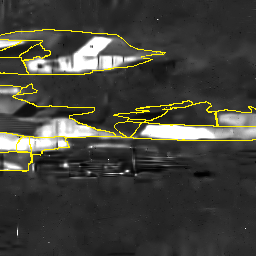}&
\includegraphics[width=4cm]{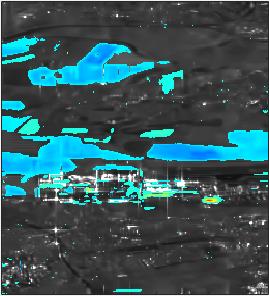}&
\includegraphics[width=4cm]{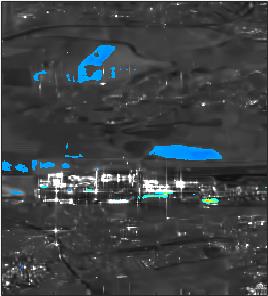}\\
\footnotesize{\rotatebox[origin=l]{90}{ 24/12/2014-05/05/2015}}&
\includegraphics[width=4.4cm]{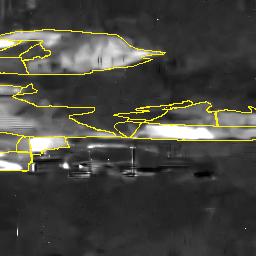}&
\includegraphics[width=4cm]{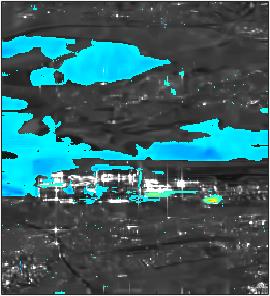}&
\includegraphics[width=4cm]{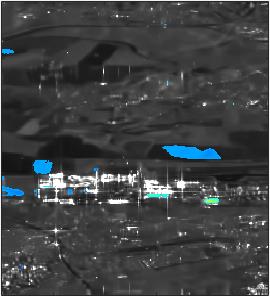}\\
\footnotesize{\rotatebox[origin=l]{90}{ 24/12/2014-05/05/2015}}&
\includegraphics[width=4.4cm]{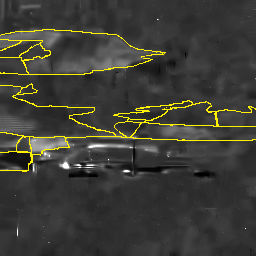}&
\includegraphics[width=4cm]{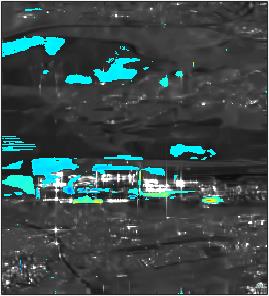}&
\includegraphics[width=4cm]{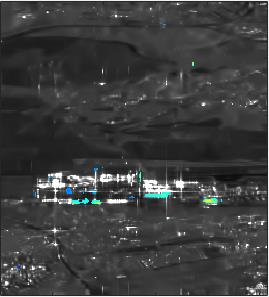}\\
\footnotesize{\rotatebox[origin=l]{90}{ 24/12/2014-05/05/2015}}&
\includegraphics[width=4.4cm]{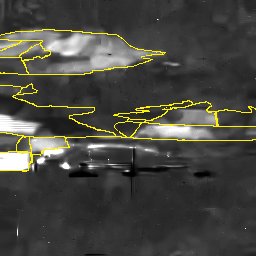}&
\includegraphics[width=4cm]{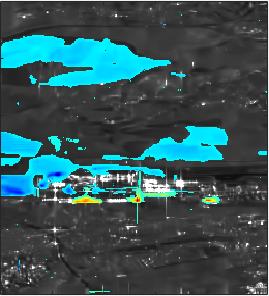}&
\includegraphics[width=4cm]{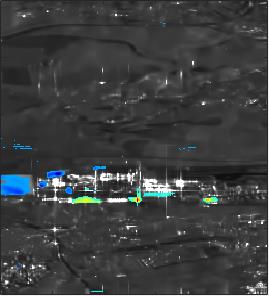}\\
&(a)&(b)&(c)\\
&\multicolumn{3}{c}{decrease \includegraphics[width=6.8cm]{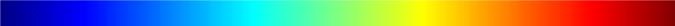} increase}\\
\\
\end{tabular}
  \caption{Continuous change monitoring results. (a) the reference image divided by the other images, (b) changes detection results, (c) changes with change magnitude weights. The thresholds were chosen for column (b) with a false alarm rate equal to 0.54\% and 1\% for column (c). The time intervals of the changes are shown on the left. Different colours  represent the decrease and increase in the magnitude of the backscattering values. The temporal images are denoised using RABASAR-DAM.}
  \label{fig:cumulativeChange} 
\end{figure}

$S_{GLR}$ method obtains the best results both with RABASAR-DAM and RABASAR-DBWAM. They all precisely detect the 5 typical changed areas (Figure \ref{fig:sendaiImgPairComparison} (c)).
However, there is some noise inside the changed areas when using RABASAR-DBWAM. $S_{GLR}$ method obtained better results for the objective change detection with RABASAR-DAM.

\subsection{Continuous change monitoring}\label{se:cumulativeChangeDetection}

Continuous change monitoring is a good way to track the development of object changes.  In this section,  5 Sentinel-1 images are processed using  $S_{GLR}$ method, which provides satisfying multitemporal change detection results.
Yellow lines are used to highlight the farmland area boundaries, as shown in the first column of Figure \ref{fig:cumulativeChange}.
Compared with the farmland backscattering values in the reference image, the others seem to have smaller values. Thus, we create the background image by dividing the reference image with other images, so as to highlight the changed farmland areas.

There are valleys in this area, and even the farmland areas are not flat. 
The threshold is defined empirically, according to the detecting change types or changed areas. For example, big threshold values lead to the detection of high change magnitude areas. With the false alarm rate equal to 0.54\% (computed using the denoised simulated SAR images without change), we can detect the farmland area changes (Figure \ref{fig:cumulativeChange}).

The positive values (red) indicate the increase of backscattering values according to the reference data, and the negative values (blue) represent the decrease of the backscattering values. The appearing or disappearing buildings always have a large change magnitude. Seasonal changed areas (like farmland areas, and some kinds of forest areas) have dynamic changes during the timeline. 
With the false alarm rate equal to 0.54\% (Figure \ref{fig:cumulativeChange}), the proposed  method detects 83.23\% of the appearing and disappearing buildings.

Since farmland areas and building areas have different mean intensity values, we could suppress the detection of farmland area changes by adding this information as a weight, with
$\exp ({\sqrt{\hat{u}_t}+\sqrt{\hat{u}_{t'}}}/{2} )S_{GLR}(\hat{u}_t,\hat{u}_{t'})$.
The exponential function is used to enlarge the backscattering value differences between different objects.

In addition, these two areas usually have different change magnitudes.
For example, with weights calculated using the log version distance of the correponding amplitude values:
\begin{equation}
    S^D_{GLR}(\hat{u}_t,\hat{u}_{t'})=\log(|\sqrt{\hat{u}_t}-\sqrt{\hat{u}_{t'}}|)S_{GLR}(\hat{u}_t,\hat{u}_{t'})
\end{equation}
we could acquire new change detection results (third column of Figure \ref{fig:cumulativeChange}). However, after multiplying the weights, the threshold has to be defined empirically. 

\subsection{Change classification}\label{se:changeClassification}
 In this section, the proposed change classification method is compared with NORCAMA \cite{su2015norcama}  with the use of   multitemporal Sentinel-1 images. 
 This process can distinguish farmland area changes from building area changes, which have seasonal and non-seasonal changes, respectively.

\subsubsection{Change classification with Sentinel-1 data}

 With high-frequency acquisition data, the whole duration of the changed buildings will be monitored. During the construction of the building, its backscattering values may keep changing which leads to  complex change monitoring results.
To suppress the complex or high-frequency changes in construction areas and keep the cycle changes of farmland areas, we under sampled the frequency of the time series. The acquisition time of the Sentinel-1 time series are  24/12/2014, 05/05/2015, 25/11/2015, 05/04/2016, 02/10/2016 and 18/01/2017.

\begin{figure}
\graphicspath{{results/}}
   \centering
\begin{tabular}{ccc}
\includegraphics[width=4cm]{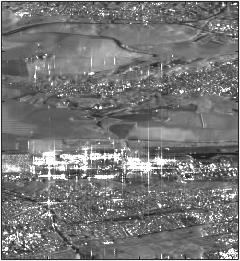}&
\includegraphics[width=4cm]{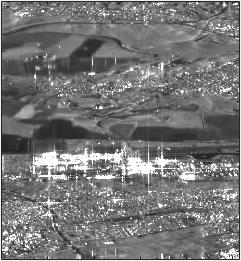}&
\includegraphics[width=4cm]{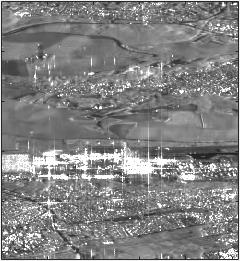}\\
 (a) 24/12/2014 &(b) 05/05/2015&(c)  25/11/2015\\
\includegraphics[width=4cm]{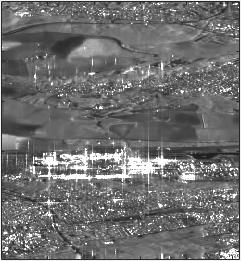}&
 \includegraphics[width=4cm]{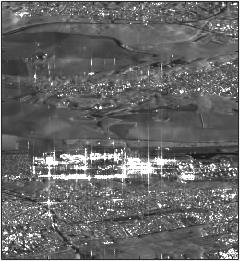}&
 \includegraphics[width=4cm]{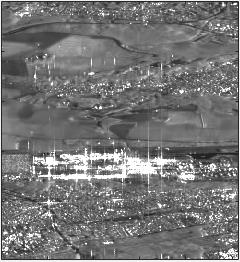}\\
 (d) 05/04/2016 &(e)  02/10/2016&(f) 18/01/2017\\
 \includegraphics[width=4cm]{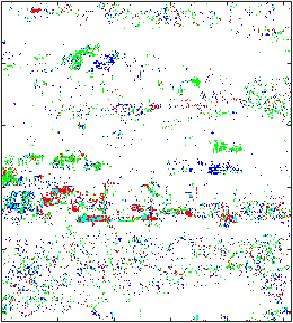}&
\includegraphics[width=4cm]{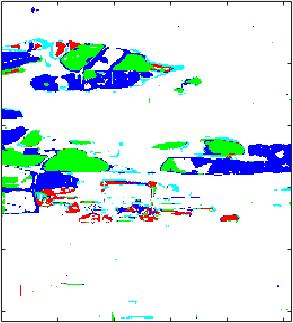}&
\includegraphics[width=4cm]{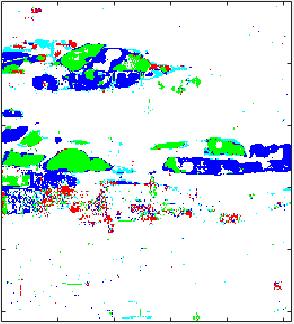}\\
 (g)&(h)&(i)\\
\end{tabular}
\caption{Sentinel-1 time series change classification. (a-f) Sentinel-1 images, (g) NORCAMA,  (h) $S_{GLR}$ with RABASAR-DAM provided data, (i) $S_{GLR}$ with RABASAR-DBWAM provided data. 6 images are used for the change type detection. The change type results are: white: no change, {red: step change}, {green: impulse change}, {blue: cycle change} and {cyan: complex change}.}
  \label{fig:Sentinel1ChangeClassification} 
\end{figure}

    However, all the Sentinel-1 IW VV polarization SAR images acquired through 110 orbit from 24/12/2014 to 18/01/2017 are used during the speckle reduction process, so as to acquire better denoising results. RABASAR-DAM denoised images and all change classification results are illustrated in Figure \ref{fig:Sentinel1ChangeClassification}. Since the actual backscattering values of the farmland areas are controlled by the surface roughness and soil moisture, we could observe the weak backscatter fields in SAR images acquired in spring. This phenomenon also reflects the seasonal changes in the time series.

Compared to NORCAMA provided results, $S_{GLR}$ based change classification method provides much better results.
All the detected changed areas are well corresponding to the previous cumulative change detection results demonstrated in Figure \ref{fig:cumulativeChange}.
Visually, $S_{GLR}$ gives better detection results (Figure \ref{fig:Sentinel1ChangeClassification}(h)) when using RABASAR-DAM provided data. There are fewer isolated points in the detection results and the changed farmland areas are very smooth. The changed types are similar to that using RABASAR-DBWAM provided data. 

Global threshold is used for $S_{GLR}$ based change type detection method, so as to speed up the processing.
The denoising results of RABASAR seem good when using denoised binary weighted arithmetic mean image, but the change type detection results are not better than that using the denoised arithmetic mean image. 
When using RABASAR provided data, NORCAMA method can obtain much better results than using 2SPPB provided data.

 \begin{figure}
\setlength{\abovecaptionskip}{0pt} 
\setlength{\belowcaptionskip}{0pt} 
\graphicspath{{results/}}
   \centering
\begin{tabular}{cc}
\includegraphics[width=5cm]{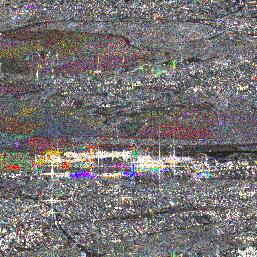}&
\includegraphics[width=5cm]{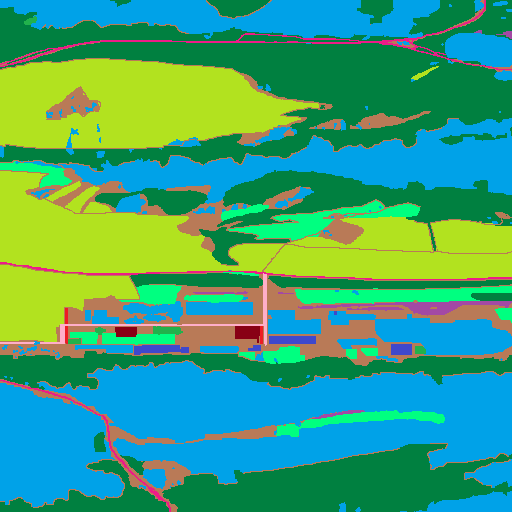}\\
(a) Maximum value time&(b) Ground truth map\\
\multicolumn{2}{c}{24/12/2014 \includegraphics[width=5.8cm]{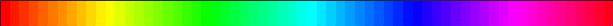} 18/01/2017}\\
\end{tabular}
  \caption{Maximum amplitude value time provided by REACTIV method. 10 Sentinel-1 images  are used for the comparison.  Ground truth map is prepared according to the arithmetic mean image, with different colours representing different objects. We mainly pay attention to the {changed building areas (blue)} and {farmland areas (yellow, green)}.}
  \label{changeTimeDetectionSent10} 
\end{figure}

\begin{figure}
\setlength{\abovecaptionskip}{0pt} 
\setlength{\belowcaptionskip}{0pt} 
\graphicspath{{results/}}
   \centering
\begin{tabular}{cc}
\includegraphics[width=5cm]{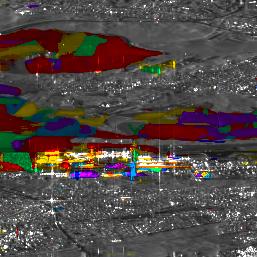}&
\includegraphics[width=5cm]{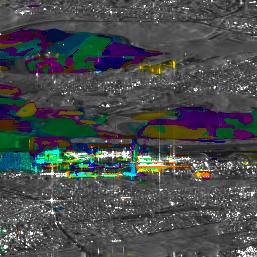}\\
(a) Maximum value time& (b) Maximum change  time\\
\includegraphics[width=5cm]{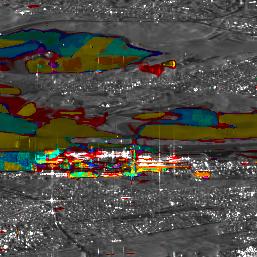}&
\includegraphics[width=5cm]{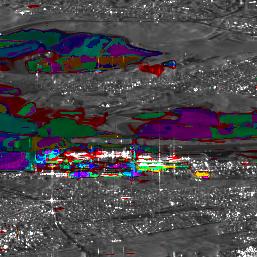}\\
(c) Start changing time&(d) Stop changing time\\
\multicolumn{2}{c}{24/12/2014 \includegraphics[width=5.8cm]{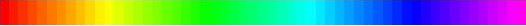} 18/01/2017}\\
\end{tabular}
  \caption{Different change time detection comparison with improved REACTIV method. 10 Sentinel-1 images are used.  The time series data are denoised by RABASAR-DAM with 69 time series images. A threshold is set at $99\%$ on the change probability. }
  \label{DiffchangeTimeDS10} 
\end{figure}

\begin{figure}
 \setlength{\abovecaptionskip}{-10pt} 
 \setlength{\belowcaptionskip}{-10pt} 
 \graphicspath{{results/}}
    \centering
 \begin{tabular}{cc}
  \multicolumn{2}{c}{
 \begin{tabular}{ccccc}
  \includegraphics[width=2.4cm]{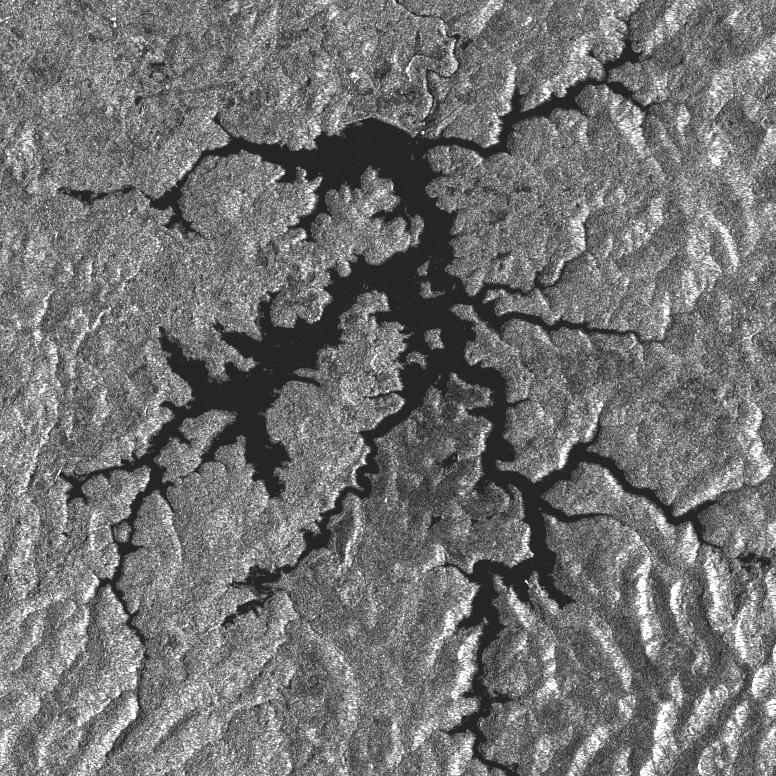}&
  \includegraphics[width=2.4cm]{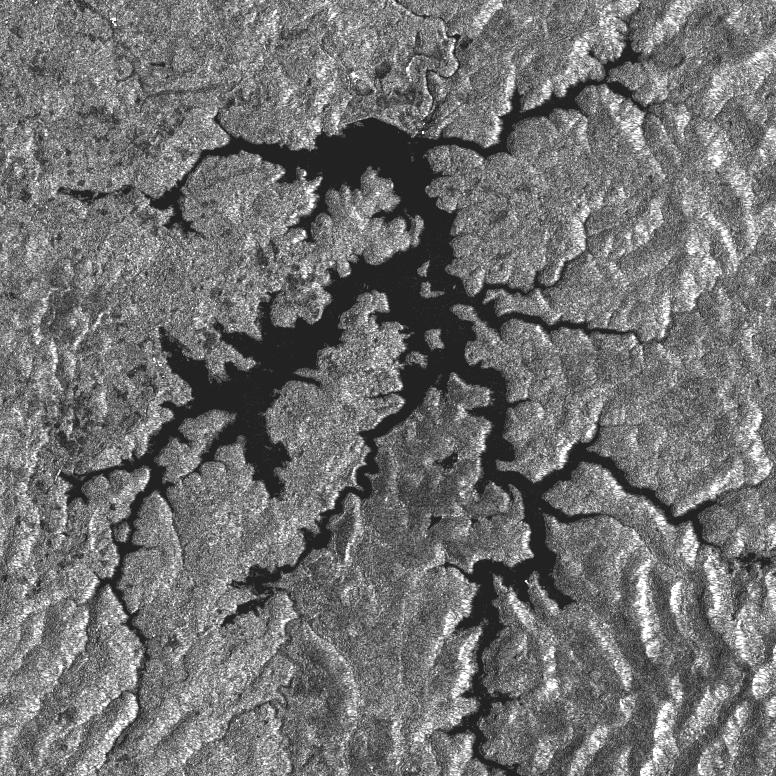}&
  \includegraphics[width=2.4cm]{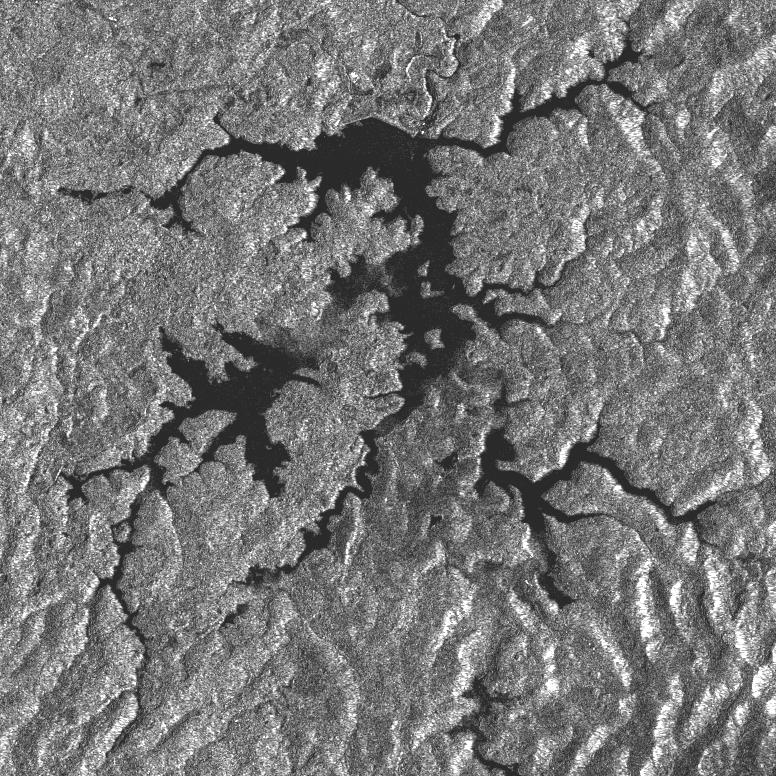}&
  \includegraphics[width=2.4cm]{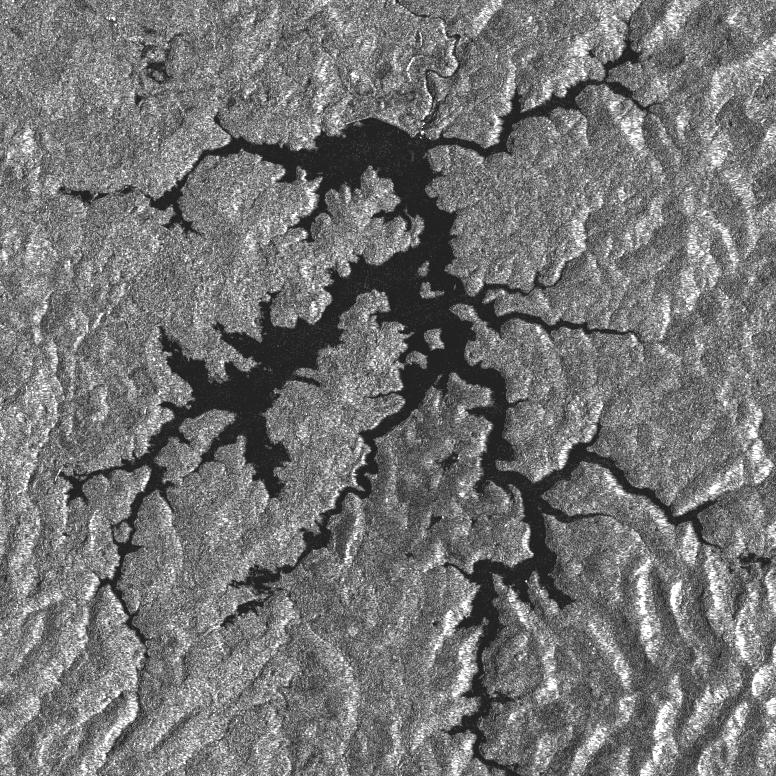}&
  \includegraphics[width=2.4cm]{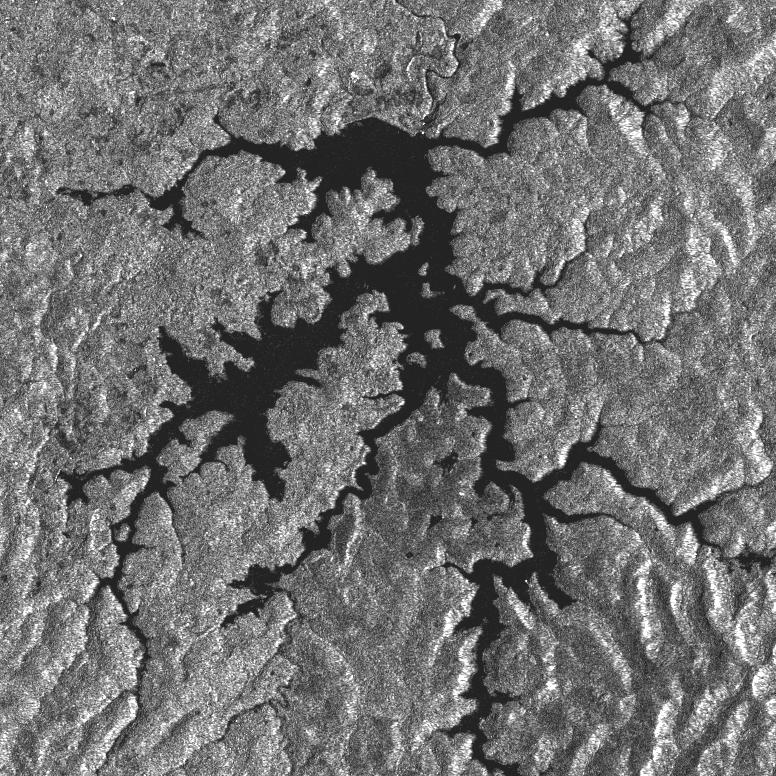}\\
 06/05/2018 & 18/05/2018 & 30/05/2018 &11/06/2018&23/06/2018\\
  \includegraphics[width=2.4cm]{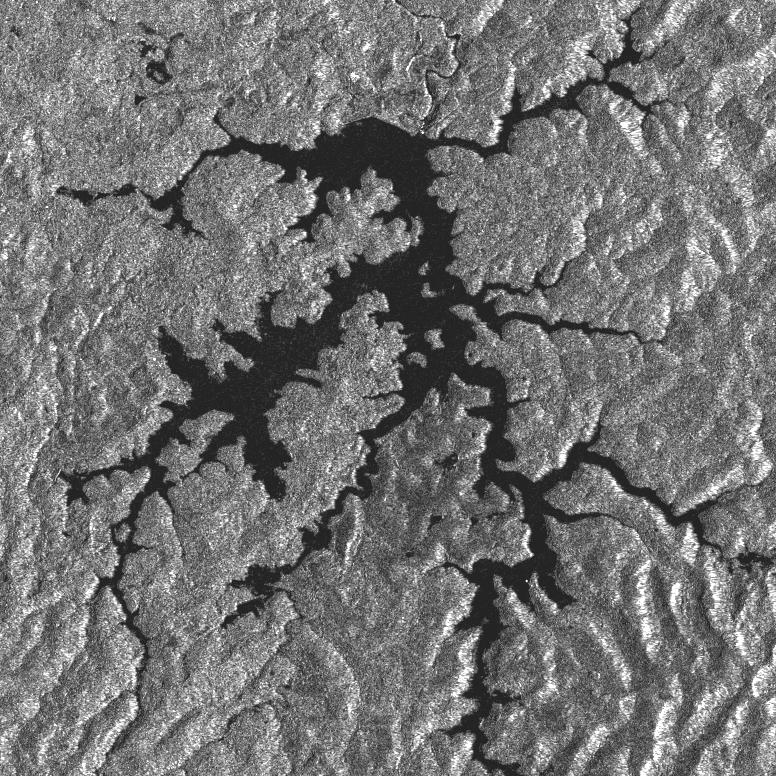}&
  \includegraphics[width=2.4cm]{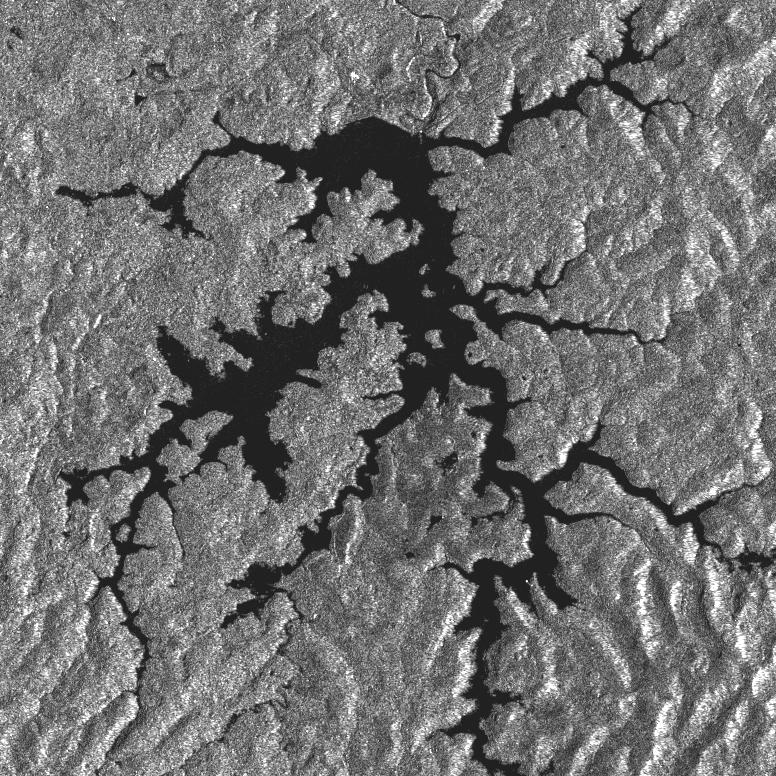}&
  \includegraphics[width=2.4cm]{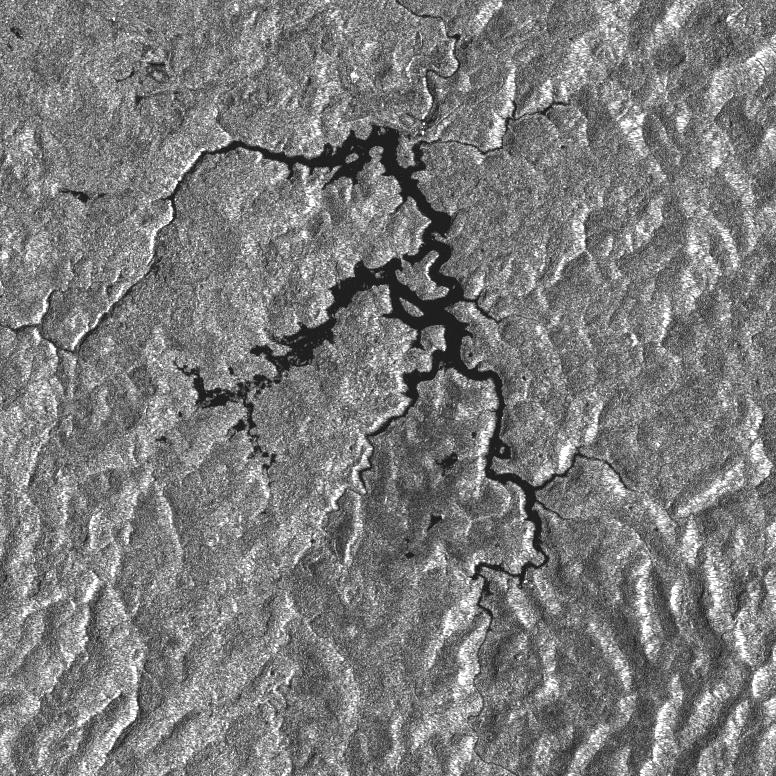}&
 \includegraphics[width=2.4cm]{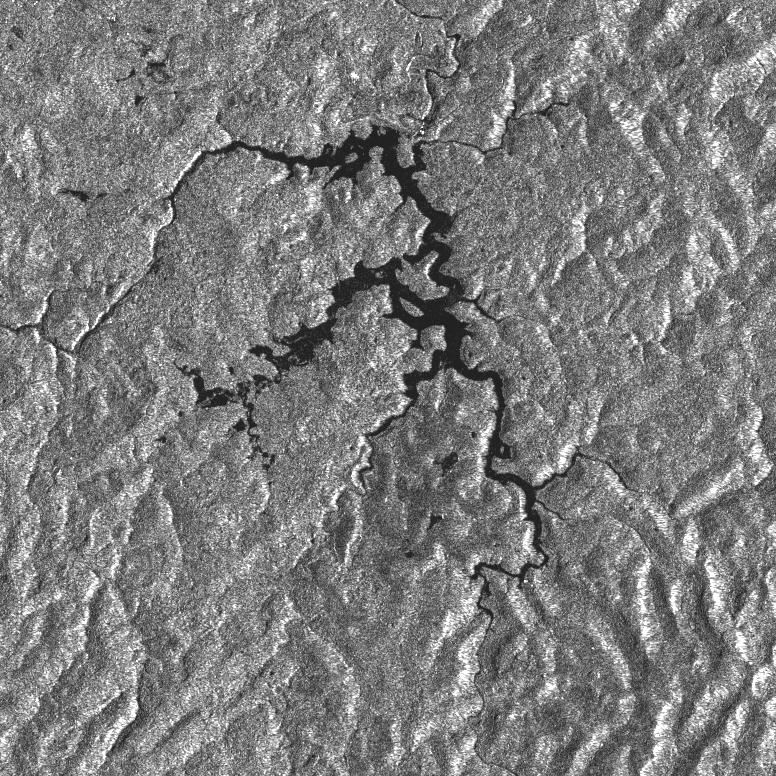}&\\
 05/07/2018 & 17/07/2018 & 29/07/2018 &10/08/2018&\\
 \end{tabular}}\\
 \includegraphics[width=6.8cm]{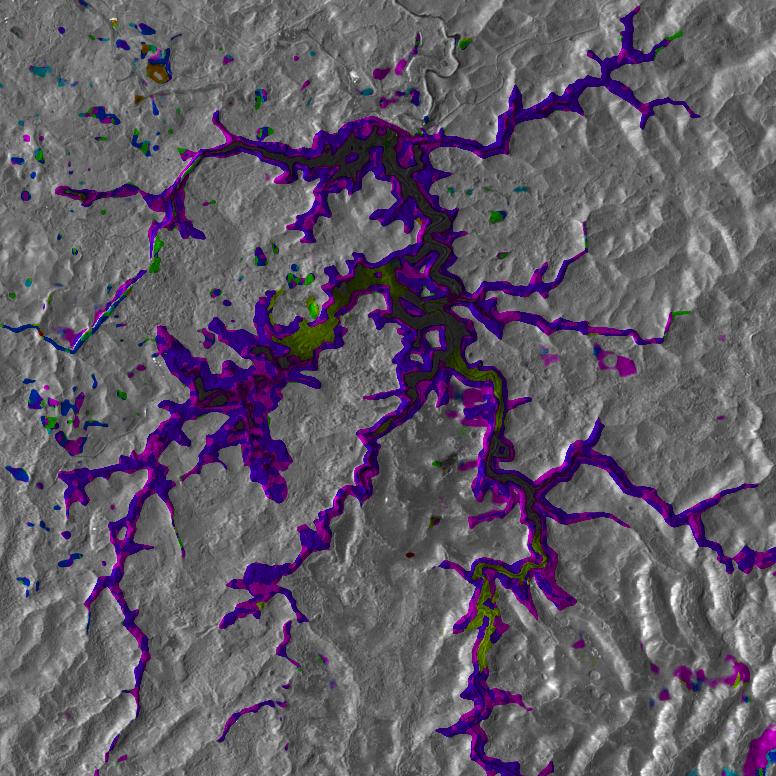}&
 \includegraphics[width=6.8cm]{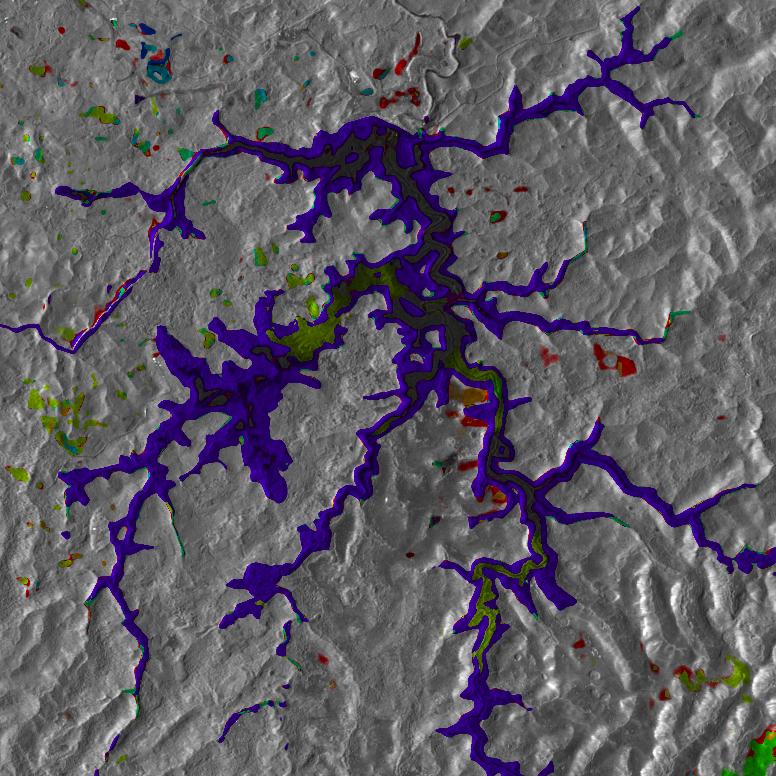}\\
 (a) Maximum value time& (b) Maximum change time\\
 \includegraphics[width=6.8cm]{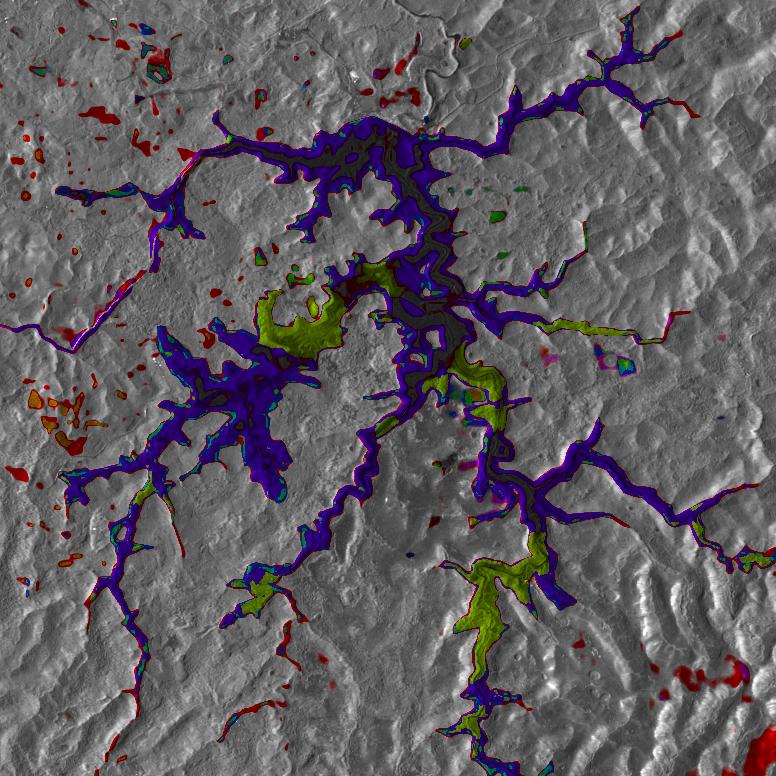}&
 \includegraphics[width=6.8cm]{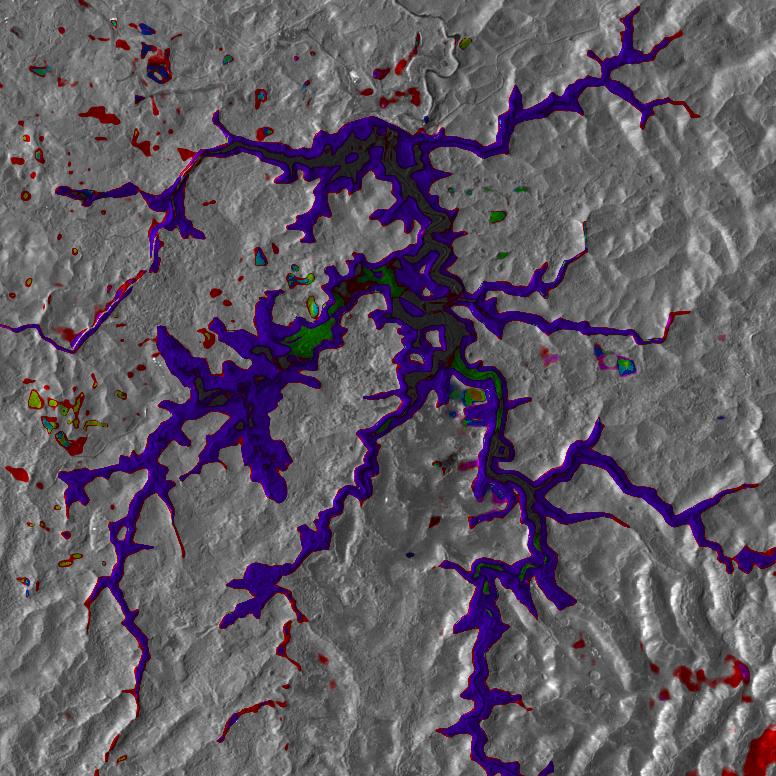}\\
 (c) Start changing time&(d) Stop changing time\\
\multicolumn{2}{c}{2018/05/06 \includegraphics[width=6.8cm]{colorbarIndex.jpeg} 2018/08/10}\\
\\
 \end{tabular}
   \caption{Flooding area change time detection comparison with 9 Sentinel-1 GRD data. 9 noisy images are shown above. The images are acquired over Xe-Pian Xe-Namnoy dam in the southeastern province of Attapeu in Laos.  87 temporal Sentinel-1 GRD images are used for the preparation of the arithmetic mean image. All the test images are provided by RABASAR-AM.}
   \label{DiffchangeTimeLaosDam} 
 \end{figure}

The same crop with similar growth periods monitored in different years with the same SAR sensor, they have similar backscattering values. It seems that using  fewer images is not enough to detect all kinds of vegetation interannual variations. 
It is recommended to use $S_{GLR}$ change type detection method with RABASAR-DAM provided data.

\subsection{Change time detection and visualization with extended REACTIV method}\label{se:changeTime}

In this section, the time series generated from 
Sentinel-1 single-look complex images and Sentinel-1 GRD images are separately used to illustrate
the capability of the proposed method.

\subsubsection{Farmland area and building area monitoring}

To make a comprehensive evaluation of the method, we processed the temporal Sentinel-1 SAR images with the original  REACTIV method. The detection results are shown in Figure \ref{changeTimeDetectionSent10}(a).
Although the changing area can be roughly detected by the test statistics with original SAR data, the results are seriously affected by the noise.
 The previously prepared ground truth map is shown in Figure \ref{changeTimeDetectionSent10}(b) to help the interpretation and evaluation of change detection results.

With the RABASAR provided data, different change time detection strategies described in Section \ref{se:ChangeTime} are utilized to process the  Sentinel-1 time series.  The spatial adaptive denoising leads to spatially variable ENL. During the change time detection test, we suppose they have the same ENL so as to speed up the process. 
In addition, the red colour at the end of the colour bar is removed so as to avoid the mix of red colours. All the time are compressed to the set of  colour index bar. 

Compared with the original method, the improved REACTIV method can obtain much better change detection results (Figure  \ref{DiffchangeTimeDS10}(a)). According to the ground truth, the improved REACTIV method can detect the changed building areas and farmland areas. With the default parameter, it can not detect the changes in forest areas which have low change magnitude in the temporal amplitude SAR images.

With reference to different object changes, the maximum change magnitude time and start changing time are the same for disappearing building areas. The detection results in the appearing building areas are much more complex.
Based on the detection results, we can even distinguish  different kinds of farmlands. Since most farmland areas are not totally flat, this may cause different parts of the same farmland to reach their maximum value at different times.
 Therefore, the detection results in these areas are not smooth enough.
Generally, construction areas are larger than the under-construction building, and this area may keep changing before the building is completed. All these phenomena  lead to complex change shapes in the detection results.

\subsubsection{Monitoring abrupt floods in Southern Laos}

Change detection is a significant application of remote sensing technology.
In this section, we try to apply the improved method to flooding area monitoring.
 There exist a large number of Sentinel-1 GRD data over the test site, all the images which have similar acquisition geometry are used to prepare the super-image.

  Sentinel-1 GRD images have the ability to monitor large-area changes. We only test the improved method over the water storage area. The amount of flooding water can be estimated according to the changes in the water area and local digital elevation model. 
 As shown in Figure \ref{DiffchangeTimeLaosDam}, most of the areas are changed between the image pairs which were acquired on 17/07/2018 and 29/07/2018.
 The black areas surrounded by blue areas are the final water area during the image acquisition period.
 The detection results have high similarity with the results provided by ESA\footnote{ESA: https://www.esa.int/Our\_Activities/Observing\_the\_Earth/Copernicus/Sentinel-1/Sentinel-1\_maps\_flash\_floods\_in\_Laos}.

 \section{Conclusion}

In this paper, we proposed a simple $S_{GLR}$ based similarity test  which could be applied and benefit to any denoised SAR images. The simple $S_{GLR}$ similarity is based on gamma distribution and used for the calculation of the criteria map.
Based on the prefiltered data, 
this method has been used for image pair change detection, continuous change monitoring and change classification. In particular, we mainly used RABASAR provided data in this paper.
The processing results of simulated and real SAR images show that $S_{GLR}$ based change detection method provided good results both in the processing of image pairs and temporal images. $S_{GLR}$ method gives better change classification results compared to NORCAMA method. Using RABASAR-DAM provided data,  $S_{GLR}$  acquired much better change classification results,  with smooth changed areas and less noisy points.

In addition, we used the $S_{GLR}$ function and RABASAR denoising data to improve REACTIV method.
Based on the detection areas acquired by REACTIV method (dynamics of time series coefficient of variation), we associated the colours with different kinds of change times and changed the background with a denoised image or arithmetic mean image. By only using part of the hue colour channel, we successfully avoid the mixture of the red colour index. The results obtained by the improved method provided useful information and allow extended interpretation.
The change time detection is much more effective for homogeneous area changes and for abrupt changes, which is suitable for monitoring farmland areas, flooding areas and some human activities (harbour activities, urbanization and airport dynamics). However, the method has less capability to detect seasonal changes in forest areas.  

Future work will take into account  the object attributes  of the changed areas, so as to acquire better analysis results. To precisely identify the seasonal change of vegetation areas, we will pay attention to the multitemporal coherence maps.



\bibliographystyle{unsrt}  
\bibliography{references}

\begin{thebibliography}{10}

\bibitem{khorram1999accuracy}
S.~Khorram.
\newblock {\em Accuracy assessment of remote sensing-derived change detection}.
\newblock Asprs Publications, 1999.

\bibitem{coppin2004review}
P.~Coppin, I.~Jonckheere, K.~Nackaerts, B.~Muys, and E.~Lambin.
\newblock Review article digital change detection methods in ecosystem
  monitoring: a review.
\newblock {\em International journal of remote sensing}, 25(9):1565--1596,
  2004.

\bibitem{habib2007abrupt}
T.~Habib, J.~Chanussot, J.~Inglada, and G.~Mercier.
\newblock Abrupt change detection on multitemporal remote sensing images: a
  statistical overview of methodologies applied on real cases.
\newblock In {\em Geoscience and Remote Sensing Symposium, 2007. IGARSS 2007.
  IEEE International}, pages 2593--2596. IEEE, 2007.

\bibitem{bruzzone2013novel}
L.~Bruzzone and F.~Bovolo.
\newblock A novel framework for the design of change-detection systems for
  very-high-resolution remote sensing images.
\newblock {\em Proceedings of the IEEE}, 101(3):609--630, 2013.

\bibitem{Lomb-02}
P.~Lombardo and T.~Pellizzeri.
\newblock Maximum likelihood signal processing techniques to detect a step
  pattern of change in multitemporal {SAR} images.
\newblock {\em IEEE Transactions on Geoscience and Remote Sensing},
  40(4):853--870, 2002.

\bibitem{su2015norcama}
X.~Su, C.A. Deledalle, F.~Tupin, and H.~Sun.
\newblock {NORCAMA}: Change analysis in {SAR} time series by likelihood ratio
  change matrix clustering.
\newblock {\em ISPRS Journal of Photogrammetry and Remote Sensing},
  101:247--261, 2015.

\bibitem{conradsen2003test}
K.~Conradsen, A.A. Nielsen, J.~Schou, and H.~Skriver.
\newblock A test statistic in the complex {W}ishart distribution and its
  application to change detection in polarimetric {SAR} data.
\newblock {\em IEEE Transactions on Geoscience and Remote Sensing},
  41(1):4--19, 2003.

\bibitem{conradsen2016determining}
K.~Conradsen, A.A. Nielsen, and H.~Skriver.
\newblock Determining the points of change in time series of polarimetric {SAR}
  data.
\newblock {\em IEEE Transactions on Geoscience and Remote Sensing},
  54(5):3007--3024, 2016.

\bibitem{PCP+16}
L.~Pulvirenti, M.~Chini, N.~Pierdicca, and G.~Boni.
\newblock Use of {SAR} data for detecting floodwater in urban and agricultural
  areas: {T}he role of the interferometric coherence.
\newblock {\em IEEE Transactions on Geoscience and Remote Sensing},
  54(3):1532--1544, 2016.

\bibitem{PMM16}
M.T. Pham, G.~Mercier, and J.~Michel.
\newblock Change detection between {SAR} images using a pointwise approach and
  graph theory.
\newblock {\em IEEE Transactions on Geoscience and Remote Sensing},
  54(4):2020--2032, 2016.

\bibitem{zhao2014deep}
J.~Zhao, M.~Gong, J.~Liu, and L.~Jiao.
\newblock Deep learning to classify difference image for image change
  detection.
\newblock In {\em Neural Networks (IJCNN), 2014 International Joint Conference
  on}, pages 411--417. IEEE, 2014.

\bibitem{gong2016change}
M.~Gong, J.~Zhao, J.~Liu, Q.~Miao, and L.~Jiao.
\newblock Change detection in synthetic aperture radar images based on deep
  neural networks.
\newblock {\em IEEE transactions on neural networks and learning systems},
  27(1):125--138, 2016.

\bibitem{TLB88}
R.~Touzi, A.~Lopes, and P.~Bousquet.
\newblock A statistical and geometrical edge detector for {SAR} images.
\newblock {\em IEEE Transactions on geoscience and remote sensing},
  26(6):764--773, 1988.

\bibitem{OQ04}
C.~Oliver and S.~Quegan.
\newblock {\em Understanding synthetic aperture radar images}.
\newblock SciTech Publishing, 2004.

\bibitem{AAD+16}
V.~Akbari, S.N. Anfinsen, A.P. Doulgeris, T.~Eltoft, G.~Moser, and S.B.
  Serpico.
\newblock Polarimetric {SAR} {C}hange {D}etection {W}ith the {C}omplex
  {H}otelling-{L}awley {T}race {S}tatistic.
\newblock {\em IEEE Transactions on Geoscience and Remote Sensing},
  54(7):3953--3966, 2016.

\bibitem{Bazi-05}
Y.~Bazi, L.~Bruzzone, and F.~Melgani.
\newblock An unsupervied approach based on the generalized {G}aussian model to
  automatic change detection in multitemporal {SAR} images.
\newblock {\em IEEE Transactions on Geoscience and Remote Sensing},
  43(4):2972--2982, 2005.

\bibitem{bovolo2005detail}
F.~Bovolo and L.~Bruzzone.
\newblock A detail-preserving scale-driven approach to change detection in
  multitemporal {SAR} images.
\newblock {\em IEEE Transactions on Geoscience and Remote Sensing},
  43(12):2963--2972, 2005.

\bibitem{bovolo2015time}
F.~Bovolo and L.~Bruzzone.
\newblock The time variable in data fusion: {A} change detection perspective.
\newblock {\em IEEE Geoscience and Remote Sensing Magazine}, 3(3):8--26, 2015.

\bibitem{aminikhanghahi2017survey}
S.~Aminikhanghahi and D.J. Cook.
\newblock A survey of methods for time series change point detection.
\newblock {\em Knowledge and information systems}, 51(2):339--367, 2017.

\bibitem{mou2018learning}
L.~Mou, L.~Bruzzone, and X.X. Zhu.
\newblock Learning spectral-spatial-temporal features via a recurrent
  convolutional neural network for change detection in multispectral imagery.
\newblock {\em arXiv preprint arXiv:1803.02642}, 2018.

\bibitem{dominguez2018multisquint}
E.M. Dom{\'\i}nguez, E.~Meier, D.~Small, M.E. Schaepman, L.~Bruzzone, and
  D.~Henke.
\newblock A multisquint framework for change detection in high-resolution
  multitemporal {SAR} images.
\newblock {\em IEEE Transactions on Geoscience and Remote Sensing},
  56(6):3611--3623, 2018.

\bibitem{nielsen2016change}
A.A. Nielsen, K.~Conradsen, and H.~Skriver.
\newblock Change detection in a short time sequence of polarimetric {C}-band
  {SAR} data.
\newblock In {\em ESA Living Planet Symposium 2016}. European Space Agency,
  ESA, 2016.

\bibitem{amitrano2016end}
D.~Amitrano, F.~Cecinati, G.~Di~Martino, A.~Iodice, P.P. Mathieu, D.~Riccio,
  and G.~Ruello.
\newblock An end-user-oriented framework for rgb representation of
  multitemporal sar images and visual data mining.
\newblock In {\em Image and Signal Processing for Remote Sensing XXII}, volume
  10004, page 100040Y. International Society for Optics and Photonics, 2016.

\bibitem{KoeniguerREACTIV}
E.~Koeniguer, A.~Boulch, P.~Trouv{\'e}, and F.~Janez.
\newblock Colored visualization of multitemporal {SAR} data for change
  detection : issues and methods.
\newblock {\em In Synthetic Aperture Radar (EUSAR), 20018 12th European
  Conference on. VDE}, 2018.

\bibitem{zhao2019ratio}
Weiying Zhao, Charles-Alban Deledalle, Lo{\"\i}c Denis, Henri Ma{\^\i}tre,
  Jean-Marie Nicolas, and Florence Tupin.
\newblock Ratio-based multitemporal sar images denoising: Rabasar.
\newblock {\em IEEE Transactions on Geoscience and Remote Sensing}, 57(6):3552
  -- 3565, 2019.

\bibitem{goodman2007speckle}
J.W. Goodman.
\newblock {\em Speckle phenomena in optics: theory and applications}.
\newblock Roberts and Company Publishers, 2007.

\bibitem{deledalle2009iterative}
C.A. Deledalle, L.~Denis, and F.~Tupin.
\newblock Iterative weighted maximum likelihood denoising with probabilistic
  patch-based weights.
\newblock {\em IEEE Transactions on Image Processing}, 18(12):2661--2672, 2009.

\bibitem{SDT+14}
X.~Su, C.A. Deledalle, F.~Tupin, and H.~Sun.
\newblock Two-step multitemporal nonlocal means for synthetic aperture radar
  images.
\newblock {\em IEEE Transactions on Geoscience and Remote Sensing},
  52(10):6181--6196, 2014.

\bibitem{wilks1938large}
S.S. Wilks.
\newblock The large-sample distribution of the likelihood ratio for testing
  composite hypotheses.
\newblock {\em The Annals of Mathematical Statistics}, 9(1):60--62, 1938.

\bibitem{anderson1958introduction}
T.W. Anderson and E.U. Math{\'e}maticien.
\newblock {\em An introduction to multivariate statistical analysis}, volume~2.
\newblock Wiley New York, 1958.

\bibitem{tison2004new}
C.~Tison, J-M. Nicolas, F.~Tupin, and H.~Ma{\^\i}tre.
\newblock A new statistical model for {M}arkovian classification of urban areas
  in high-resolution {SAR} images.
\newblock {\em IEEE transactions on geoscience and remote sensing},
  42(10):2046--2057, 2004.

\bibitem{shi2000normalized}
J.~Shi and J.~Malik.
\newblock Normalized cuts and image segmentation.
\newblock {\em IEEE Transactions on pattern analysis and machine intelligence},
  22(8):888--905, 2000.

\bibitem{ng2002spectral}
A.Y. Ng, M.I. Jordan, and Y.~Weiss.
\newblock On spectral clustering: Analysis and an algorithm.
\newblock In {\em Advances in neural information processing systems}, pages
  849--856, 2002.

\bibitem{schubert2014signitrend}
E.~Schubert, M.~Weiler, and H.P. Kriegel.
\newblock Signitrend: scalable detection of emerging topics in textual streams
  by hashed significance thresholds.
\newblock In {\em Proceedings of the 20th ACM SIGKDD international conference
  on Knowledge discovery and data mining}, pages 871--880. ACM, 2014.

\bibitem{von2007tutorial}
U.~Von~Luxburg.
\newblock A tutorial on spectral clustering.
\newblock {\em Statistics and computing}, 17(4):395--416, 2007.

\bibitem{koeniguervisualisation}
E.C. Koeniguer, J-M. Nicolas, B.~Pinel-Puyssegur, J-M. Lagrange, and F.~Janez.
\newblock Visualisation des changements sur s{\'e}ries temporelles radar:
  m{\'e}thode {REACTIV} {\'e}valu{\'e}e {\`a} l'{\'e}chelle mondiale sous
  {G}oogle {E}arth {E}ngine.
\newblock {\em Conf{\'e}rence Fran\c{c}aise de Photogramm{\'e}trie et de
  T{\'e}l{\'e}d{\'e}tection (CFPT)}, 2018.

\bibitem{basseville1993detection}
Mich{\`e}le Basseville, Igor~V Nikiforov, et~al.
\newblock {\em Detection of abrupt changes: theory and application}, volume
  104.
\newblock Prentice Hall Englewood Cliffs, 1993.

\bibitem{G76}
J.W. Goodman.
\newblock Some fundamental properties of speckle.
\newblock {\em Journal of the Optical Society of America}, 66(11):1145--1150,
  1976.

\bibitem{nicolas2006application}
J-M. Nicolas.
\newblock Application de la transform{\'e}e de {M}ellin: {\'e}tude des lois
  statistiques de l’imagerie coh{\'e}rente.
\newblock {\em Rapport de recherche, 2006D010}, 2006.

\bibitem{kendall1977advanced}
M.~Kendall and A.~Stuart.
\newblock The advanced theory of statistics. {V}ol. 1: {D}istribution theory.
\newblock {\em London: Griffin, 1977, 4th ed.}, 1977.

\bibitem{JERFFO12}
J-M. Nicolas, E.~Trouve, R.~Fallourd, F.~Vernier, F.~Tupin, O.~Harant, M.~Gay,
  and L.~Moreau.
\newblock A first comparison of {C}osmo-{S}kymed and {T}erra{SAR}-{X} data over
  {C}hamonix {M}ont-{B}lanc test-site.
\newblock {\em In Geoscience and Remote Sensing Symposium (IGARSS)}, pages
  5586--5589, July 2012.

\bibitem{quin2014mimosa}
G.~Quin, B.~Pinel-Puyssegur, J-M. Nicolas, and P.~Loreaux.
\newblock {MIMOSA}: {A}n automatic change detection method for {SAR} time
  series.
\newblock {\em IEEE Transactions on Geoscience and Remote Sensing},
  52(9):5349--5363, 2014.

\end{thebibliography}

\end{document}